\documentclass{article} 
\usepackage{iclr2027,times}

\usepackage{amsmath,amsfonts,bm}

\def\eqref#1{equation~\ref{#1}}

\def\1{\bm{1}}

\DeclareMathAlphabet{\mathsfit}{\encodingdefault}{\sfdefault}{m}{sl}
\SetMathAlphabet{\mathsfit}{bold}{\encodingdefault}{\sfdefault}{bx}{n}

\usepackage{hyperref}
\usepackage{url}
\usepackage{booktabs}
\usepackage{multirow}
\usepackage{tabularx}
\usepackage{array}
\usepackage{makecell}
\usepackage{graphicx}
\usepackage{subcaption}
\usepackage{wrapfig}
\usepackage{capt-of}
\usepackage{needspace}
\usepackage{placeins}

\newcolumntype{Y}{>{\centering\arraybackslash}X}

\usepackage{xspace}
\newcommand{\modelname}{\textsc{ActTime}\xspace}

\title{ActTime: Action-Conditioned World Modeling for Time Series}

\author{
Xinxing Zhou$^{1}$,
Xiaojie Yuan$^{1}$,
Yanlong Wen$^{1*}$,
Ming Jin$^{2}$\thanks{\hspace{-0.5em}$^{*}$Corresponding authors.} \\
$^{1}$Nankai University \\
$^{2}$Griffith University \\
\texttt{zhouxinxing@dbis.nankai.edu.cn}, \\
\texttt{wenyl@nankai.edu.cn},
\texttt{mingjinedu@gmail.com}
}

\iclrfinalcopy 
\begin{document}

\maketitle
\fancyhead[L]{Preprint}
\begin{abstract}
Time-series models increasingly condition on known future inputs.
In controlled systems, some of these inputs are actions that can be
changed, compared, and optimized before execution.
A model must therefore do more than predict the realized future:
it must predict how outcomes change under different actions from the
same observed history and support reliable comparison of candidate plans.
We introduce \modelname{}, an action-conditioned world model for time series.
Over the observed history, \modelname{} updates variable-specific latent states
through an action-conditioned transition and corrects them using observations.
It then rolls the same transition forward under candidate actions to generate
future trajectories.
Training uses only factual trajectories, without paired alternative-action
outcomes or planning supervision.
We construct a simulator-grounded evaluation suite across building energy,
traffic, greenhouse climate, and inventory management, covering factual
prediction, intervention-effect prediction, offline decision making, and
closed-loop planning.
\modelname{} ranks first in 12 of 16 domain-task settings.
Across eight methods, the mean pairwise model-ranking reversal rate
is 17.8\% between factual prediction and intervention fidelity,
and rises to 28.4\% between intervention fidelity and decision ranking.
These results show that factual accuracy, intervention fidelity,
and decision quality are related but not interchangeable. 
\end{abstract}

\section{Introduction}
\label{sec:Introduction}

Modern time-series forecasting increasingly incorporates information beyond the
target history. Models use multivariate observations, static attributes,
exogenous covariates, and known-future inputs
\citep{lim2021temporal,das2023longterm,wang2024timexer,li2026gcgnet,han2026unica}.
These advances broaden forecasting from history-only extrapolation to
conditional prediction with rich contextual information \citep{yamaguchi2026citras,meyer2026t0timeseriesfoundationmodel}.
In controlled systems, however, some future inputs serve as decision variables
rather than fixed context: they are actions selected by a decision maker and
can be changed before execution, compared against alternatives, and optimized
for downstream objectives.
A forecasting model can in principle condition on an action sequence, but
decision making requires more than predicting the future under one prescribed
sequence.
Given the same observed history, the model must predict how future outcomes
vary across candidate action sequences, rather than only the outcome under
the realized sequence, since these relative differences determine which
actions should be preferred before execution.

\begin{figure*}[t]
    \centering
    \begin{subfigure}{0.32\textwidth}
        \centering
        \includegraphics[width=\textwidth]{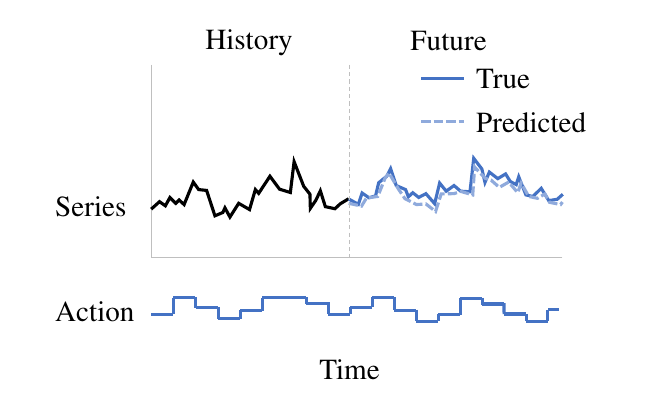}
        \caption{Factual prediction}
        \label{fig:1a}
    \end{subfigure}
    \begin{subfigure}{0.32\textwidth}
        \centering
        \includegraphics[width=\textwidth]{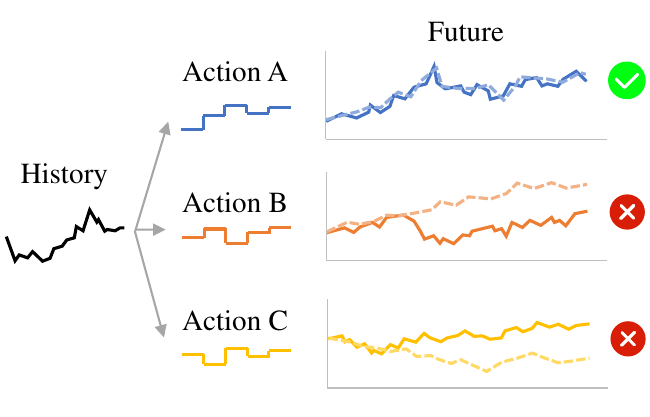}
        \caption{Intervention-effect prediction}
        \label{fig:1b}
    \end{subfigure}
    \begin{subfigure}{0.32\textwidth}
        \centering
        \includegraphics[width=\textwidth]{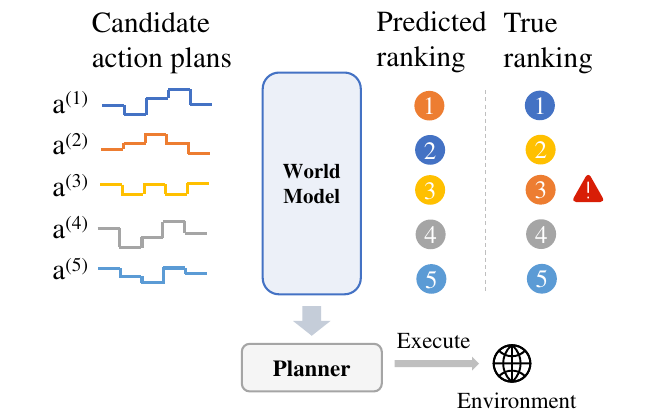}
        \caption{Decision ranking}
        \label{fig:1c}
    \end{subfigure}
    \caption{From factual prediction to decision making.
    Accurate prediction under observed actions \textbf{(a)} does not guarantee faithful responses under alternative actions \textbf{(b)}.
    Errors in these responses can further distort candidate-action rankings
    and lead to suboptimal decisions \textbf{(c)}.}
    \label{fig:1}
\end{figure*}

Figure~\ref{fig:1} makes the resulting evaluation gaps explicit.
Factual evaluation assesses the future generated by the action sequence that
was actually executed, as in Fig.~\ref{fig:1a}.
Accurate prediction of this trajectory does not directly test whether changing
the future actions produces the correct change in the resulting trajectory.
This gives the first gap:
\textbf{factual accuracy does not guarantee intervention fidelity}.
A model can fit the realized future well while still misestimating the effect of an
alternative action sequence, as illustrated in Fig.~\ref{fig:1b}.
A similar failure mode has been observed in recent action-conditioned world
models, where performance on demonstrated actions can mask incorrect responses
to valid alternative actions \citep{wang2026thermoforce,chen2026robotic}.
The second gap appears when these predictions are used to choose actions.
A planner compares candidate action sequences according to their predicted
outcomes, so the decision depends on their \emph{relative ordering}, not only
on the accuracy of each individual response.
Prediction errors can therefore reverse the ordering of candidate actions even
when their trajectory-level errors appear moderate \citep{donti2017task,elmachtoub2022smart}.
Thus, \textbf{intervention fidelity does not by itself guarantee decision
quality}, as illustrated in Fig.~\ref{fig:1c}.
Related work on world-model evaluation has similarly moved beyond predictive
or visual fidelity toward direct assessment of policy rankings and closed-loop
task performance
\citep{quevedo2026worldgym,zhang2026world,yu2026should}.
Our experiments quantify both gaps across four controlled domains and eight
methods: the average pairwise method-ordering reversal rate is $17.8\%$ from
factual prediction to intervention fidelity and rises to $28.4\%$ from
intervention fidelity to decision ranking.

These gaps call for a model that represents controlled dynamics and can be
rolled forward under alternative actions.
This is the central role of a world model: learning action-conditioned dynamics
that can be queried under candidate actions before they are executed
\citep{hafner2019learning,hafner2020iclr-dream,hansen2024td}.
While this view is well established in model-based control and embodied agents,
related ideas have also been explored in action-conditioned numerical dynamics
and, more recently, in controlled and interventional time-series modeling
\citep{shaj2021action,shaj2023multi,wang2026thermoforce,meng2026cedar,
wu2026doflow}.
Yet a factual trajectory reveals only the action sequence that was executed and
the future that followed it; matched outcomes under alternative actions are not
observed.
This leaves a practical question:
can a time-series model learn
\textbf{action-conditioned dynamics from factual trajectories that remain
reliable under alternative actions for intervention-effect prediction and decision
making}?

We address this question with \modelname, an action-conditioned time-series
world model built around a shared controlled transition.
Over the observed history, variable-specific latent states are propagated under
the executed actions and corrected using each new observation, allowing the
transition to model action-conditioned evolution while the latent state remains
grounded in the realized trajectory.
At the history boundary, a lightweight residual prior from a frozen pretrained
temporal backbone refines the terminal state.
The same transition is then rolled forward under candidate future actions
without access to future observations.
Alternative-action predictions therefore arise directly from the action-conditioned dynamics learned on factual trajectories.
To evaluate this capability, we construct simulator-grounded evaluation suites
across four controlled systems.
Training uses only factual trajectories, while same-history interventions,
candidate-action comparisons, and closed-loop planning episodes are generated
only for evaluation.
This provides direct ground truth for the progression in Fig.~\ref{fig:1}
without paired alternative-action outcomes or planning supervision during training.

Our main contributions are threefold:
\begin{itemize}
    \item
    \textbf{Decision-oriented formulation and evaluation gaps.}
    We distinguish factual accuracy, intervention fidelity, and decision
    quality as three distinct evaluation criteria for action-conditioned
    time-series models, and show that model performance does not reliably
    transfer across adjacent criteria.

    \item
    \textbf{Action-conditioned time-series world model.}
    We introduce \modelname, which learns variable-specific latent states with
    a shared action-conditioned transition and observation-based correction.
    The same transition is used for observed-history modeling and open-loop
    rollout under candidate future actions, using factual supervision only.

    \item
    \textbf{Simulator-grounded evaluation suite.}
    We construct evaluation suites across four heterogeneous domains that cover
    factual prediction, same-history intervention fidelity, offline decision making, and closed-loop planning.
    The underlying simulators provide reference action effects and decision
    outcomes, enabling direct evaluation of whether action-conditioned models
    remain reliable when used for decision making.
\end{itemize}

 \section{Related Work}
\label{sec:Related_Works}

\paragraph{Time-Series Forecasting with Covariates}

Time-series forecasting has long incorporated information beyond target history, including static attributes, exogenous variables, and known-future inputs \citep{wen2017multi,rangapuram2018deep,salinas2020deepar}. NBEATSx, TFT, and TiDE integrate such covariates through dedicated components or feature selection \citep{olivares2023neural,lim2021temporal,das2023longterm}. More recent models focus on interactions between endogenous and exogenous series, including TimeXer, ExoTST, CrossLinear, and GCGNet \citep{wang2024timexer,tayal2024exotst,zhou2025crosslinear,li2026gcgnet}. Covariate-aware forecasting has also reached TSFMs \citep{yamaguchi2026citras}. Chronos-2 supports multivariate and covariate-informed forecasting \citep{ansari2025chronos}, while UniCA adapts pretrained TSFMs to heterogeneous covariates \citep{han2026unica}. These models show that contextual information can improve forecasting, but an action is not merely given context. It is changed, compared, and optimized. Factual accuracy therefore provides limited evidence about responses to alternative actions.

\paragraph{Action-Conditioned Time-Series Modeling}

Several recent studies explicitly model actions or interventions in temporal systems. ThermoForce separates passive dynamics from control-induced responses and combines a frozen TSFM prior with a structured intervention model for HVAC control \citep{wang2026thermoforce}. CEDAR learns action-conditioned demand dynamics for counterfactual simulation and budget planning \citep{meng2026cedar}. DRIFT and NormWear-2 model physiological trajectories under treatment interventions \citep{liu2026drift,luo2026toward}, while DoFlow unifies observational, interventional, and counterfactual time-series prediction in a causal generative framework \citep{wu2026doflow}. Related work in longitudinal causal inference estimates outcomes under alternative treatment sequences using recurrent, representation-learning, and Transformer-based models \citep{lim2018forecasting,Bica2020Estimating,melnychuk2022causal,xiong2024g}. These methods mainly target domain-specific intervention modeling or causal treatment-effect estimation. We instead study a common action-conditioned modeling framework across heterogeneous controlled systems and evaluate whether predictive accuracy translates into intervention fidelity and decision quality.

\paragraph{World Models for Decision Making}

World models learn dynamics that can be rolled forward under candidate actions for planning and policy learning \citep{pmlr-v235-ma24i,pmlr-v267-zhou25t}. PlaNet and Dreamer use latent imagined trajectories for decision making \citep{hafner2019learning,hafner2020iclr-dream,hafner2025mastering}, while TD-MPC2 combines latent dynamics with trajectory optimization \citep{hansen2024td}. AcRKN and MTS3 model action-conditioned transitions in numerical dynamical systems, including long-horizon prediction under control inputs \citep{shaj2021action,shaj2023multi}. Another line studies reusable dynamics priors. APV pretrains from action-free videos before learning action-conditioned dynamics for control \citep{seo2022reinforcement}, with related ideas explored in recent world-model pretraining \citep{pmlr-v267-gao25u,wan2026learning}. Evaluation has also moved beyond predictive fidelity. WorldGym tests whether simulated rollouts preserve policy performance and rankings \citep{quevedo2026worldgym}, World-In-World evaluates closed-loop task success \citep{zhang2026world}, and recent work examines fidelity under alternative actions \citep{chen2026robotic}. We follow this decision-oriented view and evaluate factual prediction, intervention fidelity, offline decision making, and closed-loop planning.

\section{Methodology}
\label{sec:Methodology}

\paragraph{Problem Formulation.}
We consider multivariate time-series systems whose future evolution depends on
control actions.
Let $\mathbf{o}_t \in \mathbb{R}^{D}$ denote the observation at time step $t$
and $\mathbf{a}_t \in \mathcal{A} \subseteq \mathbb{R}^{A}$ the corresponding
action.
We denote the target variables by
$\mathbf{y}_t=\mathbf{o}_t[\mathcal{I}]\in\mathbb{R}^{P}$,
where $\mathcal{I}$ indexes the $P$ variables of interest.
We index each action with the observation obtained after its execution.

Given an observed history $\mathbf{o}_{1:T}$ and $\mathbf{a}_{1:T}$,
together with a candidate future action sequence
$\mathbf{a}_{T+1:T+H}$, the model predicts
\begin{equation}
\hat{\mathbf{y}}_{T+1:T+H}
=
f_{\Theta}\!\left(
\mathbf{o}_{1:T},
\mathbf{a}_{1:T},
\mathbf{a}_{T+1:T+H}
\right).
\label{eq:problem}
\end{equation}

For a fixed observed history, the model can be queried with multiple candidate
future action sequences
$\{\mathbf{a}^{(k)}_{T+1:T+H}\}_{k=1}^{K}$
to obtain their corresponding predicted trajectories
$\{\hat{\mathbf{y}}^{(k)}_{T+1:T+H}\}_{k=1}^{K}$.
Comparing a candidate trajectory with a designated reference trajectory
characterizes its predicted intervention effect, while applying a downstream
objective to these trajectories enables candidate-action ranking and planning.

Training uses only factual trajectories, where each history is paired with one
realized future action sequence and its resulting target trajectory.
No same-history alternative futures are used for training.

\begin{figure}[t]
    \centering
    \includegraphics[width=\textwidth]{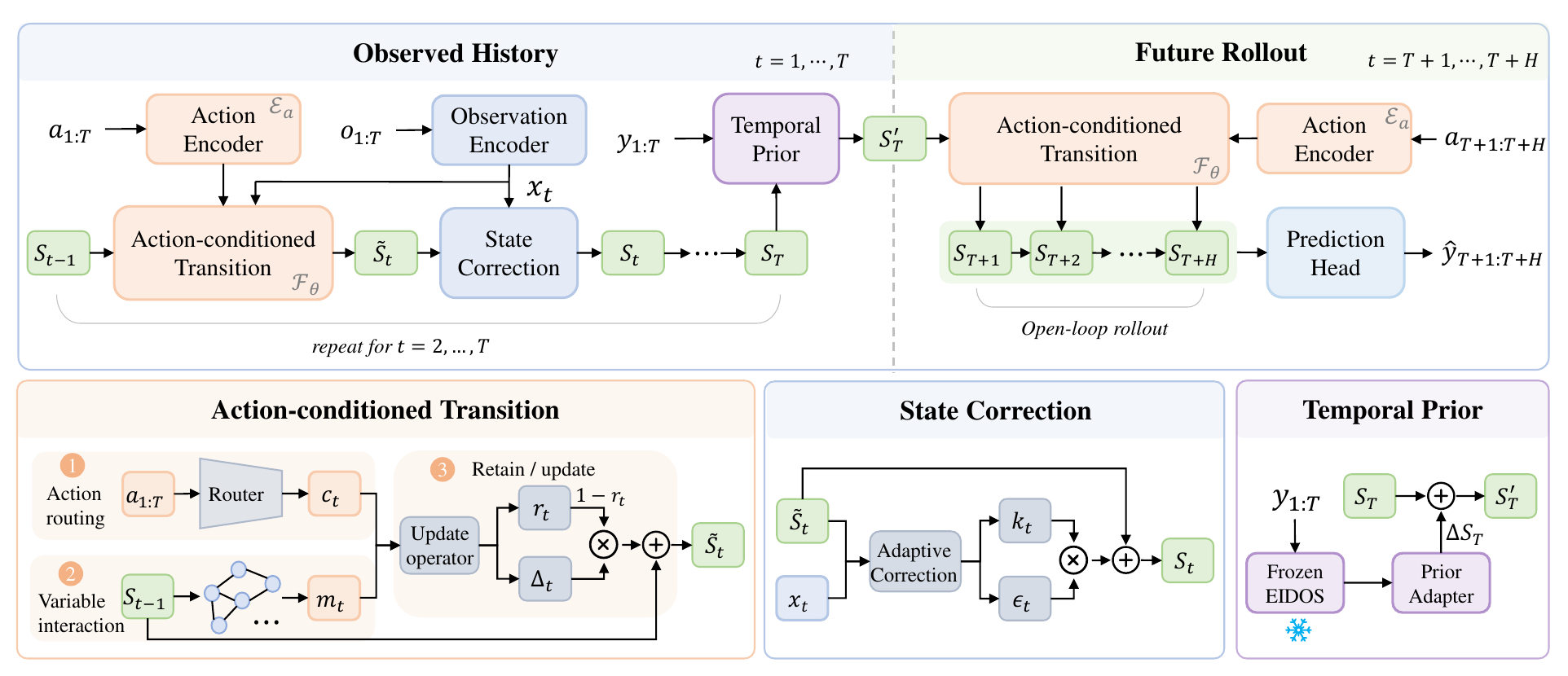}
    \caption{Overview of \modelname.
    The model filters the observed history by alternating a shared
    action-conditioned transition with observation-based state correction.
    A frozen pretrained temporal backbone provides a history prior through a
    lightweight residual adapter, after which the same transition rolls the
    state forward under candidate future actions for prediction.
    The bottom panels detail the transition, correction, and prior modules.}
    \label{fig:archi}
\end{figure}

\paragraph{Overview.}
Figure~\ref{fig:archi} summarizes \modelname.
Over the observed history, variable-specific latent states are propagated by
an action-conditioned transition and corrected using each new observation.
At the history boundary, a frozen pretrained temporal backbone provides a
history prior through a lightweight residual adapter.
The resulting terminal state is then rolled forward by the same transition
under candidate future actions without access to future observations, and a
prediction head decodes the simulated states into target quantiles.

\subsection{Action-Conditioned State Dynamics}
\label{sec:state_dynamics}

We represent the system at time $t$ by
$\mathbf{S}_t\in\mathbb{R}^{D\times d_s}$, which maintains a latent state for
each of the $D$ observed variables, with state dimension $d_s$.
Learned variable embeddings
$\mathbf{V}\in\mathbb{R}^{D\times d_s}$ provide persistent variable identity.

The observation encoder represents each current value together with its local
change, while the action encoder maps the full action vector and its change
to a control representation for each variable:
\begin{equation}
\mathbf{x}_t
=
\mathcal{E}_o([\mathbf{o}_t,\Delta\mathbf{o}_t]),
\qquad
\mathbf{c}_t
=
\mathcal{E}_a([\mathbf{a}_t,\Delta\mathbf{a}_t]),
\label{eq:state_inputs}
\end{equation}
where $[\cdot,\cdot]$ denotes concatenation,
$\Delta\mathbf{o}_t=\mathbf{o}_t-\mathbf{o}_{t-1}$,
$\Delta\mathbf{a}_t=\mathbf{a}_t-\mathbf{a}_{t-1}$, and
$\mathbf{x}_t,\mathbf{c}_t\in\mathbb{R}^{D\times d_s}$.
The observation encoder is shared across variables, whereas
$\mathcal{E}_a$ uses all action channels to produce a separate control
representation for each variable state.
The differences are set to zero at $t=1$, and the initial state is
\begin{equation}
\mathbf{S}_1
=
\mathcal{E}_{\mathrm{init}}
\left(
\mathbf{x}_1+\mathbf{c}_1+\mathbf{V}
\right),
\label{eq:state_init}
\end{equation}
where $\mathcal{E}_{\mathrm{init}}$ is a learned normalization and projection.

\paragraph{Action-conditioned transition.}
The transition consists of a stack of blocks that share the same structure.
For clarity, we describe one block below and omit the block index.
Given the incoming state, we first exchange information across variables
through self-attention:
\begin{equation}
\bar{\mathbf{S}}_{t-1}
=
\operatorname{LN}(\mathbf{S}_{t-1}),
\qquad
\mathbf{m}_t
=
\sigma(\alpha)\,
\operatorname{MHA}
\left(
\bar{\mathbf{S}}_{t-1},
\bar{\mathbf{S}}_{t-1},
\bar{\mathbf{S}}_{t-1}
\right),
\label{eq:variable_interaction}
\end{equation}
where $\alpha$ is a learned scalar controlling the contribution of the
cross-variable message.
The normalized state, interaction message, and routed control are then
concatenated as
$\mathbf{h}_t=
[\bar{\mathbf{S}}_{t-1},\mathbf{m}_t,\mathbf{c}_t]$.
A bounded retention gate and a candidate state update are computed from this
joint representation:
\begin{equation}
\mathbf{r}_t
=
g_r(\mathbf{h}_t),
\qquad
\boldsymbol{\Delta}_t
=
\operatorname{SwiGLU}
\left(
\operatorname{SiLU}(W_\Delta\mathbf{h}_t+b_\Delta)
\right),
\label{eq:transition_update}
\end{equation}
where $g_r$ is a learned sigmoid-gated projection and
$\mathbf{r}_t\in(0,1)^{D\times d_s}$.
Each block updates the state according to
\begin{equation}
\mathcal{T}
(\mathbf{S}_{t-1},\mathbf{c}_t)
=
\operatorname{LN}\!\left(
\mathbf{S}_{t-1}
+
(1-\mathbf{r}_t)\odot\boldsymbol{\Delta}_t
\right).
\label{eq:transition_block}
\end{equation}
Applying the transition blocks sequentially defines the shared
action-conditioned transition $F_\theta$.
The update therefore depends jointly on the previous state, interactions among
variables, and the applied action, allowing action effects to propagate through
the latent dynamics rather than entering only at the prediction stage.

\paragraph{State correction.}
The transition predicts how the latent state evolves under the applied action.
During the observed history, this prediction is corrected using the current
observation representation.
With
$\mathbf{z}_t=
[\operatorname{LN}(\tilde{\mathbf{S}}_t),\mathbf{x}_t]$,
where
$\tilde{\mathbf{S}}_t=F_\theta(\mathbf{S}_{t-1},\mathbf{c}_t)$,
the correction module computes
\begin{equation}
\boldsymbol{\epsilon}_t
=
\tanh(W_\epsilon\mathbf{z}_t+b_\epsilon),
\qquad
\mathbf{k}_t
=
\sigma(W_k\mathbf{z}_t+b_k),
\qquad
\mathbf{S}_t
=
\tilde{\mathbf{S}}_t
+
\mathbf{k}_t\odot\boldsymbol{\epsilon}_t .
\label{eq:state_correction}
\end{equation}
Here, $\boldsymbol{\epsilon}_t$ is the observation-dependent state update and
$\mathbf{k}_t\in(0,1)^{D\times d_s}$ controls its contribution.
The transition models action-conditioned evolution, while the correction
incorporates the realized observations.
Iterating these operations over the observed history produces the terminal
state $\mathbf{S}_T$.

\subsection{Temporal Prior and Future Rollout}
\label{sec:prior_rollout}

\paragraph{Temporal prior.}
At the history boundary, \modelname augments $\mathbf{S}_T$ with temporal
information from a frozen pretrained EIDOS \citep{zhou2026eidos} backbone.
Each target history in $\mathbf{y}_{1:T}$ is standardized over time and
encoded independently.
The resulting features are passed through a frozen projection to $d_p$ dimensions and averaged
across targets, and the final-time representation
$\mathbf{p}_T\in\mathbb{R}^{d_p}$ is used as the temporal prior.
After broadcasting $\mathbf{p}_T$ across the $D$ variable states, a
lightweight residual adapter computes
\begin{equation}
\Delta\mathbf{S}_T
=
\gamma\,\tanh\!\left(
\operatorname{MLP}
[
\operatorname{RMSNorm}(\mathbf{p}_T),\mathbf{V}
]
\right),
\qquad
\mathbf{S}'_T
=
\mathbf{S}_T+\Delta\mathbf{S}_T .
\label{eq:temporal_prior}
\end{equation}
The scalar gate $\gamma$ is initialized to zero.
The prior is applied only at the history boundary, while the parameters of the
action-conditioned transition remain unchanged.

\paragraph{Future rollout and prediction.}
Starting from $\mathbf{S}'_T$, the same transition $F_\theta$ is used to
simulate the future under the candidate action sequence.
Future actions are encoded by $\mathcal{E}_a$ as in
Eq.~\ref{eq:state_inputs}, with the first action difference computed relative
to the last observed action $\mathbf{a}_T$:
\begin{equation}
\mathbf{S}_{T+1}
=
F_\theta(\mathbf{S}'_T,\mathbf{c}_{T+1}),
\qquad
\mathbf{S}_{T+h}
=
F_\theta(\mathbf{S}_{T+h-1},\mathbf{c}_{T+h}),
\quad h=2,\ldots,H.
\label{eq:future_rollout}
\end{equation}
No observation correction is applied during rollout because future
observations are unavailable.
The future trajectory is therefore generated solely from the inferred history
state, the candidate actions, and the learned controlled dynamics.

To decode each future state, we select the latent states corresponding to the
target indices and summarize the full system state by
\begin{equation}
\mathbf{Z}_{T+h}
=
\mathbf{S}_{T+h}[\mathcal{I}]
\in\mathbb{R}^{P\times d_s},
\qquad
\bar{\mathbf{s}}_{T+h}
=
\operatorname{Mean}_{D}(\mathbf{S}_{T+h})
\in\mathbb{R}^{d_s}.
\label{eq:readout_context}
\end{equation}
The prediction head combines each target state with the shared context
$\bar{\mathbf{s}}_{T+h}$ and outputs
$\boldsymbol{\delta}_{T+h}$,
$\boldsymbol{\ell}_{T+h}$, and
$\boldsymbol{u}_{T+h}\in\mathbb{R}^{P}$.
Using the last observed target $\mathbf{y}_T$ as an anchor, the prediction
head produces ordered quantiles as
\begin{equation}
\begin{aligned}
\hat{\mathbf{y}}^{\,0.5}_{T+h}
&=
\mathbf{y}_T+\boldsymbol{\delta}_{T+h},\\
\hat{\mathbf{y}}^{\,0.1}_{T+h}
&=
\hat{\mathbf{y}}^{\,0.5}_{T+h}
-\operatorname{softplus}(\boldsymbol{\ell}_{T+h}),
\qquad
\hat{\mathbf{y}}^{\,0.9}_{T+h}
=
\hat{\mathbf{y}}^{\,0.5}_{T+h}
+\operatorname{softplus}(\boldsymbol{u}_{T+h}).
\end{aligned}
\label{eq:quantile_head}
\end{equation}
This parameterization guarantees ordered quantiles, and
$\hat{\mathbf{y}}^{\,0.5}_{T+h}$ is used as the point prediction.

For a fixed observed history, all candidate future action sequences start from
the same terminal state $\mathbf{S}'_T$.
Rolling this state forward under different candidates yields their
corresponding future trajectories, enabling same-history comparison of alternative actions and
candidate-plan evaluation.

\subsection{Model Training}
\label{sec:training}

We train \modelname using factual trajectories with a quantile prediction
objective.
Let $\mathcal{Q}=\{0.1,0.5,0.9\}$ denote the predicted quantile levels.
For a future target $\mathbf{y}_{T+h}$ and its quantile prediction
$\hat{\mathbf{y}}_{T+h}^{\,q}$, the prediction loss is
\begin{equation}
\mathcal{L}_{\mathrm{pred}}
=
\frac{1}{HP|\mathcal{Q}|}
\sum_{h=1}^{H}
\sum_{j=1}^{P}
\sum_{q\in\mathcal{Q}}
\rho_q
\left(
y_{T+h}^{(j)}-\hat{y}_{T+h}^{\,q,(j)}
\right),
\qquad
\rho_q(u)=\max\{qu,(q-1)u\}.
\label{eq:training_loss}
\end{equation}

We separate learning the controlled dynamics from adapting the pretrained
temporal prior.
First, the temporal-prior pathway is disabled and
Eq.~\ref{eq:training_loss} is used to train the observation and action
encoders, shared controlled dynamics, state correction, and prediction head.
This yields the factual action-conditioned predictor used as the controlled
dynamics core.

We then freeze the entire learned predictor and the EIDOS-based prior
extractor, and optimize only the lightweight residual adapter in
Eq.~\ref{eq:temporal_prior}.
Because its scalar gate is initialized at zero,
\begin{equation}
\gamma=0
\quad\Longrightarrow\quad
\Delta\mathbf{S}_T=\mathbf{0},
\qquad
\mathbf{S}'_T=\mathbf{S}_T,
\label{eq:zero_gate}
\end{equation}
the second optimization stage begins with exactly the same predictions as the
first-stage model.
Adaptation is therefore restricted to a residual modification of the terminal
history state, while the learned controlled dynamics and prediction head remain
fixed.

Both stages optimize the same factual prediction objective in
Eq.~\ref{eq:training_loss}.
No same-history alternative-action outcomes or planning objectives are used in
either stage.

\section{Experiments}
\label{sec:experiments}

\subsection{Experimental Setup}
\label{sec:experimental_setup}

\paragraph{Environments and data.}
We evaluate on four simulation environments:
CityLearn for building energy management
\citep{nweye2023citylearn},
SUMO Cologne1 for traffic-signal control
\citep{ault2021reinforcement,lopez2018microscopic},
GreenLight for greenhouse climate control
\citep{van2025greenlight,katzin2020greenlight},
and gym-invmgmt for serial inventory management
\citep{barati2026gym}.
We generate all training and evaluation trajectories directly from the
corresponding simulators.
Training contains one realized future per history.
For intervention-effect and offline decision evaluation, each held-out history is
paired with multiple future action plans whose outcomes are obtained from the
simulator.
Online evaluation executes selected actions in the simulator and replans from
the resulting observations.
Alternative-action outcomes are not used for training or final checkpoint
selection.
Table~\ref{tab:domains} summarizes the domain configurations.

\begin{table}[t]
\centering
\caption{Simulation environments and task configurations.
$D/A/P$ denote the observation, action, and target dimensions, and $T/H$
the history/prediction horizons in simulator steps.}
\label{tab:domains}
\small
\renewcommand{\arraystretch}{1.12}
\setlength{\tabcolsep}{3.5pt}
\begin{tabularx}{\linewidth}{
@{}l
c
c
c
>{\hsize=0.80\hsize\raggedright\arraybackslash}X
>{\hsize=1.20\hsize\raggedright\arraybackslash}X
@{}}
\toprule
Domain & $D/A/P$ & Step & $T/H$ & Action & Targets \\
\midrule
CityLearn
& 28/1/2 & 1 h & 24/12
& Battery charge/discharge
& SOC, net electricity consumption \\
SUMO
& 25/4/6 & 10 s & 24/12
& Traffic-signal phase
& Queue, waiting time, traffic flow \\
GreenLight
& 36/6/3 & 15 min & 96/16
& Climate actuators
& CO$_2$, temperature, humidity \\
Inventory
& 15/3/4 & 1 period & 24/12
& Replenishment requests
& Backlog, echelon inventories \\
\bottomrule
\end{tabularx}
\end{table}

\paragraph{Evaluation.}
We evaluate factual prediction, intervention fidelity, offline decision making,
and closed-loop planning.
Factual prediction uses MAE in target-wise standardized space.
Intervention fidelity uses Effect NMAE, which measures error in the predicted
alternative-versus-reference trajectory effect relative to the magnitude of
the true effect.
Offline decision making reports the macro average of normalized regret over
three predefined objectives, using the simulator oracle and uniform candidate
selection as references.
Closed-loop planning reports each domain's primary
control objective under receding-horizon model predictive control (MPC).
Lower is better throughout.
Exact metric and objective definitions are given in the Appendix~\ref{app:evaluation}.

\paragraph{Baselines.}
We compare with five forecasting baselines:
Ridge-VARX, DLinear~\citep{zeng2023transformers},
PatchTST~\citep{DBLP:conf/iclr/NieNSK23},
iTransformer~\citep{liu2024itransformer}, and
TFT~\citep{lim2021temporal},
together with two latent world models,
RSSM~\citep{hafner2019learning} and
MTS3~\citep{shaj2023multi}.
All methods receive historical actions and candidate future actions, with each
prediction conditioned only on actions available up to that horizon.
The same held-out histories and candidate action banks are used across methods.
Baseline adaptations are provided in the Appendix~\ref{app:baselines}.

\subsection{Main Results}
\label{sec:main_results}

Table~\ref{tab:main_results} summarizes performance across four domains and
four evaluation settings.
\modelname ranks first in 12 of the 16 task-domain settings and achieves the
best overall average rank of 1.50, compared with 2.25 for the strongest
baseline. These results span continuous and discrete control settings,
with action spaces ranging from scalar battery control and traffic-signal phases
to multichannel climate actuation and replenishment decisions.

\begin{table}[t]
\centering
\caption{
Main results across four controlled dynamical systems and four evaluation
settings.
Lower is better for all task metrics ($\downarrow$).
Macro Regret denotes the average normalized regret over the three predefined
offline objectives.
Best and second-best results are shown in \textbf{bold} and
\underline{underlined}, respectively.
Ridge denotes Ridge-VARX, while DLinear, PatchTST, iTransformer, and TFT
denote their action-conditioned variants.
}
\label{tab:main_results}

\scriptsize
\renewcommand{\arraystretch}{1.08}
\setlength{\tabcolsep}{2.5pt}

\begin{tabularx}{\textwidth}{
@{}
p{1.30cm}
p{1.75cm}
*{8}{Y}
@{}
}
\toprule
\textbf{Domain}
&
\textbf{Task / Metric}
&
\textbf{Ridge}
&
\textbf{DLinear}
&
\textbf{PatchTST}
&
\textbf{iTransformer}
&
\textbf{TFT}
&
\textbf{RSSM}
&
\textbf{MTS3}
&
\textbf{\modelname}
\\
\midrule

\multirow{4}{*}{CityLearn}
&
Factual MAE
&
0.2875
&
0.3585
&
0.3410
&
0.3602
&
0.2583
&
\underline{0.2501}
&
0.2512
&
\textbf{0.2486}
\\

&
Effect NMAE
&
0.1495
&
0.3493
&
0.2700
&
0.3972
&
\textbf{0.0498}
&
0.0790
&
\underline{0.0716}
&
0.0725
\\

&
Macro Regret
&
0.4793
&
0.3869
&
0.4396
&
0.5050
&
\underline{0.2880}
&
0.3739
&
0.3105
&
\textbf{0.2798}
\\

&
Online Objective
&
0.8310
&
0.8541
&
0.8481
&
0.9099
&
\underline{0.8250}
&
\textbf{0.8196}
&
0.8414
&
0.8317
\\

\midrule

\multirow{4}{*}{SUMO}
&
Factual MAE
&
0.3873
&
0.4358
&
0.4150
&
0.3730
&
\underline{0.1226}
&
0.1730
&
0.2134
&
\textbf{0.1076}
\\

&
Effect NMAE
&
1.0851
&
0.9201
&
0.9202
&
0.9060
&
\textbf{0.4184}
&
0.4914
&
0.5776
&
\underline{0.4825}
\\

&
Macro Regret
&
0.5255
&
0.5661
&
0.5662
&
0.5620
&
\underline{0.0296}
&
0.0802
&
0.0759
&
\textbf{0.0221}
\\

&
Online Objective
&
0.3940
&
0.3834
&
0.3836
&
0.3843
&
\underline{0.3147}
&
0.3193
&
0.3430
&
\textbf{0.3036}
\\

\midrule

\multirow{4}{*}{GreenLight}
&
Factual MAE
&
0.3263
&
0.3699
&
0.3647
&
0.3398
&
\underline{0.0666}
&
0.1008
&
0.1242
&
\textbf{0.0498}
\\

&
Effect NMAE
&
0.8622
&
0.9384
&
0.9221
&
0.8860
&
\underline{0.2811}
&
0.3761
&
0.5367
&
\textbf{0.2091}
\\

&
Macro Regret
&
0.4352
&
0.7962
&
0.8965
&
0.7096
&
0.2367
&
0.1956
&
\underline{0.1870}
&
\textbf{0.1397}
\\

&
Online Objective
&
0.3105
&
0.3201
&
0.3232
&
0.3258
&
\underline{0.2946}
&
0.3009
&
0.2961
&
\textbf{0.2944}
\\

\midrule

\multirow{4}{*}{Inventory}
&
Factual MAE
&
0.2210
&
0.3812
&
0.3938
&
0.3706
&
\underline{0.0992}
&
0.1243
&
0.1279
&
\textbf{0.0790}
\\

&
Effect NMAE
&
0.6538
&
0.9711
&
0.9774
&
0.9567
&
\underline{0.3077}
&
0.3506
&
0.4008
&
\textbf{0.2244}
\\

&
Macro Regret
&
\textbf{0.1169}
&
0.2929
&
0.3221
&
0.3133
&
0.1585
&
0.2130
&
\underline{0.1531}
&
0.1563
\\

&
Online Objective
&
4.1079
&
4.2951
&
4.3760
&
4.2875
&
\underline{3.8917}
&
3.9583
&
4.0445
&
\textbf{3.8739}
\\

\midrule

\multirow{2}{*}{\textbf{Summary}}
&
\textbf{\# Best $\uparrow$}
&
1
&
0
&
0
&
0
&
\underline{2}
&
1
&
0
&
\textbf{12}
\\

&
\textbf{Avg. Rank $\downarrow$}
&
5.19
&
6.75
&
7.06
&
6.63
&
\underline{2.25}
&
3.19
&
3.44
&
\textbf{1.50}
\\

\bottomrule
\end{tabularx}
\end{table}

\textbf{Factual prediction.}
\modelname achieves the lowest factual MAE in all four domains.
Relative to the strongest baseline, MAE is reduced by 12.2\% on SUMO,
25.2\% on GreenLight, and 20.4\% on Inventory.
Thus, the gains under alternative actions do not come at the cost of accuracy
on the realized trajectories used in conventional forecasting evaluation.

\textbf{Intervention fidelity.}
The factual ordering does not carry over unchanged once future actions are
varied.
TFT performs best on CityLearn and SUMO, while \modelname leads on GreenLight
and Inventory with 25.6\% and 27.1\% lower Effect NMAE than the strongest
baseline.
This shift indicates that factual accuracy alone does not determine how well a
model captures the effects of alternative actions.
Section~\ref{sec:alignment} examines this mismatch directly.

\textbf{Decision making.}
The most consistent gains appear when predictions are used to choose actions.
\modelname achieves the lowest Offline Macro Regret on CityLearn, SUMO, and
GreenLight, and the best closed-loop objective on SUMO, GreenLight, and
Inventory.
Its offline regret is about 25\% lower than the strongest baseline on both
SUMO and GreenLight.
The decision-level ordering also differs from intervention fidelity.
For example, \modelname ranks second on SUMO Effect NMAE but first in both
offline and closed-loop decision evaluation.
This provides an initial indication that accurate effect prediction and useful
decision rankings, while related, are not equivalent.

\subsection{Alignment Across Evaluation Levels}
\label{sec:alignment}

The rank changes in Table~\ref{tab:main_results} motivate a direct test of
cross-level alignment.
We compare adjacent evaluation levels on matched held-out histories.
Prediction measures reference-trajectory error, Effect measures the fidelity of
action-induced trajectory changes, and Decision measures candidate-action
ranking quality.
For each method pair, we record whether its relative ordering reverses between
adjacent levels.
Using orderings rather than raw metric differences makes the comparison
insensitive to the different scales of the three criteria.
Detailed definitions and per-domain statistics are provided in the Appendix~\ref{app:cross_level_alignment}.

\Needspace{0.30\textheight}
\begin{wrapfigure}[13]{l}{0.6\textwidth}
    \centering
    \vspace{-0.7em}
    \includegraphics[
        width=\linewidth
    ]{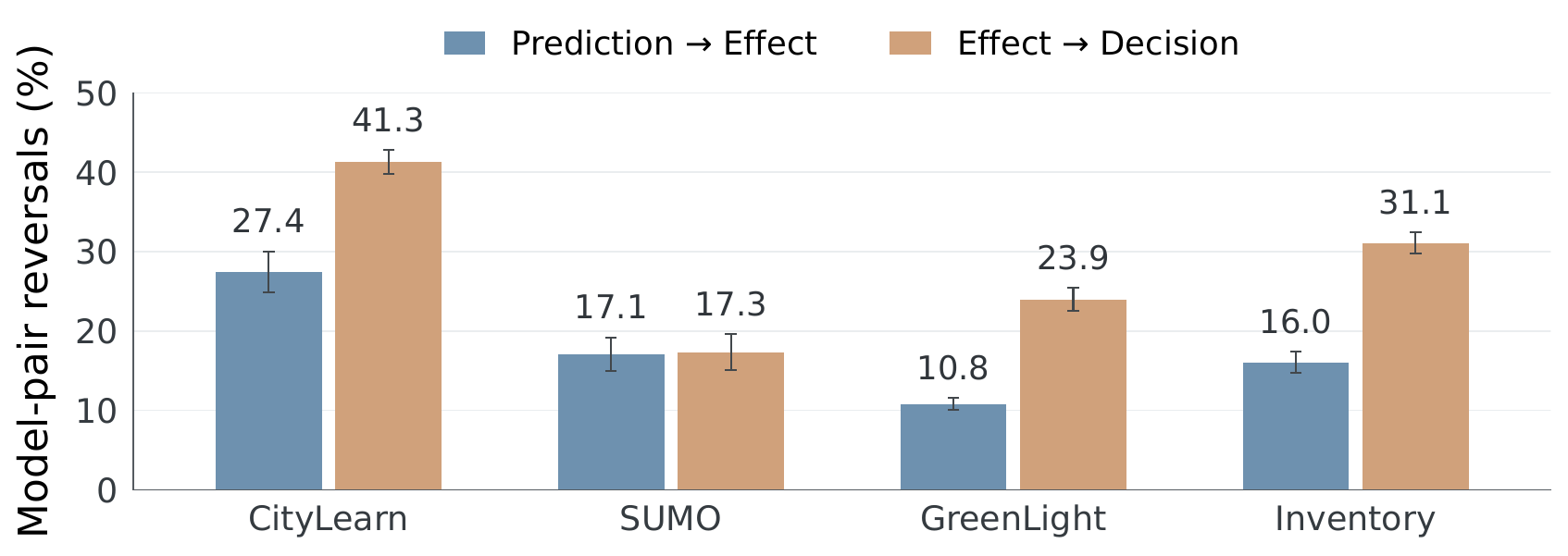}
    \caption{
    Model-order reversals across evaluation levels.
    Bars report the fraction of method pairs whose ordering reverses between
    adjacent levels.
    Error bars denote 95\% bootstrap confidence intervals over histories.
    }
    \label{fig:metric_alignment}
    \vspace{-0.8em}
\end{wrapfigure}

Figure~\ref{fig:metric_alignment} shows weaker alignment toward the decision level.
The mean reversal rate increases from 17.8\% for Prediction-to-Effect to
28.4\% for Effect-to-Decision, while the corresponding macro Spearman
correlation decreases from 0.763 to 0.514.
The increase in reversals appears in all four domains, with the largest
Effect-to-Decision gaps on CityLearn and Inventory.

Upstream and downstream performance remain positively associated in aggregate,
yet this association is not strong enough to preserve the ordering of
individual methods.
An upstream metric can therefore remain informative while still being an
unreliable criterion for selecting the better model downstream.
For Effect-to-Decision, reversal rates range from 17.3\% on SUMO to 41.3\%
on CityLearn.
Factual accuracy, intervention fidelity, and decision quality should therefore
be treated as distinct but related evaluation levels.

\subsection{Candidate-Action Ranking Fidelity}
\label{sec:ranking_fidelity}

The preceding analysis shows that intervention fidelity does not fully characterize
decision quality. A planner selects among candidate actions according to their
relative predicted costs, so we next evaluate whether this ordering is preserved
directly. For each held-out history, we compute the cost of every candidate plan
from the model-predicted trajectory and compare it with the corresponding
simulator cost under the domain's predefined primary objective. We report the
Spearman rank correlation averaged over histories, where higher values indicate
more faithful candidate-action ordering.

\begin{table}[t]
    \centering
    \caption{
    Candidate-action ranking fidelity.
    Mean per-history Spearman correlation between predicted and simulator
    candidate costs under each domain's predefined primary decision objective.
    Higher is better. Best and second-best results are shown in bold and
    underlined, respectively.
    }
    \label{tab:ranking_fidelity}
    \small
    \setlength{\tabcolsep}{4.5pt}
    \begin{tabular}{lcccccccc}
        \toprule
        Domain
        & Ridge
        & DLinear
        & PatchTST
        & iTransformer
        & TFT
        & RSSM
        & MTS3
        & \modelname{} \\
        \midrule
        CityLearn
        & \underline{0.8608}
        & 0.8213
        & 0.8361
        & 0.8053
        & 0.8381
        & 0.7627
        & 0.8123
        & \textbf{0.8810} \\

        SUMO
        & 0.4745
        & 0.4709
        & 0.4639
        & 0.4619
        & \underline{0.9008}
        & 0.8532
        & 0.8451
        & \textbf{0.9048} \\

        GreenLight
        & 0.5164
        & 0.2612
        & 0.1970
        & 0.2713
        & \underline{0.7527}
        & 0.7284
        & 0.7334
        & \textbf{0.8514} \\

        Inventory
        & 0.7934
        & 0.6064
        & 0.5799
        & 0.6026
        & 0.7703
        & 0.7757
        & \underline{0.8071}
        & \textbf{0.8271} \\
        \midrule
        Avg.
        & 0.6613
        & 0.5399
        & 0.5192
        & 0.5352
        & \underline{0.8154}
        & 0.7800
        & 0.7995
        & \textbf{0.8661} \\
        \bottomrule
    \end{tabular}
\end{table}

Table~\ref{tab:ranking_fidelity} shows that \modelname{} achieves the highest
candidate-ranking correlation in all four domains. Its macro-average
correlation is 0.8661, compared with 0.8154 for TFT, the strongest baseline on
this metric. The advantage is especially pronounced on GreenLight and remains
consistent across CityLearn, SUMO, and Inventory. Together with the offline regret results in Table~\ref{tab:main_results},
this shows that the decision performance of \modelname{} is accompanied by
more faithful ordering of the full candidate set, rather than being limited
to the ultimately selected plan.

We further test whether this result depends on the choice of decision objective
or ranking metric. Across all 12 domain-objective combinations,
\modelname{} achieves the best macro performance under Spearman correlation,
Kendall's $\tau_b$, and pairwise ordering accuracy. It ranks first on 11 of the
12 objectives under Spearman correlation and on 10 of 12 under both Kendall's
$\tau_b$ and pairwise accuracy. Detailed results are reported in
Appendix~\ref{sec:ranking_appendix}.

\subsection{Ablation Study}
\label{sec:ablation}

We ablate the main components of the controlled dynamics.
Because metric scales differ across domains, we normalize each ablation by the
corresponding full-model metric within each domain and then macro-average the
ratios.
Table~\ref{tab:ablation} reports the resulting values, where values above
$1.0$ indicate degradation.

\begin{table}[t]
    \centering
    \caption{
    Ablation study averaged across the four domains.
    Each entry is the ratio to the matched full-model metric within each domain,
    so $1.0$ denotes parity and larger values indicate degradation.
    }
    \label{tab:ablation}
    \small
    \renewcommand{\arraystretch}{1.08}
    \setlength{\tabcolsep}{5pt}
    \begin{tabular}{lcccc}
        \toprule
        Variant
        & Factual MAE
        & Effect NMAE
        & Macro Regret
        & Online Objective \\
        \midrule

        \textbf{Full}
        & 1.000
        & 1.000
        & 1.000
        & 1.000 \\

        Simple Action Concatenation
        & 1.059
        & 1.035
        & 1.137
        & 1.021 \\

        w/o State Correction
        & 2.137
        & 2.132
        & 2.215
        & 1.031 \\

        Shared State Across Variables
        & 5.787
        & 5.367
        & 4.360
        & 1.146 \\

        w/o Action Conditioning
        & 4.290
        & 6.393
        & 16.349
        & 2.999 \\

        \bottomrule
    \end{tabular}
\end{table}

The two action ablations separate the value of action information from the way
it enters the dynamics.
Removing action conditioning produces the largest degradation on intervention
and decision metrics, increasing Effect NMAE by more than sixfold and Offline
Macro Regret by more than sixteenfold.
Replacing the structured action pathway with simple concatenation is less
destructive, but still degrades every metric, with the clearest loss in offline
decision making.
Action information is therefore essential, while the structured pathway provides a better overall balance across evaluation criteria than simple action concatenation.

The state representation is similarly important.
Collapsing the variable-specific latent states into a shared state severely
degrades factual prediction, intervention fidelity, and offline decision
making.
Removing state correction roughly doubles these three errors.
Together, these results support maintaining heterogeneous system states and
re-grounding them with observations before open-loop rollout.
The weaker degradation in the online objective is consistent with
receding-horizon replanning repeatedly incorporating fresh simulator
observations, which limits the accumulation of open-loop model error.
Per-domain structural ablations and an exactly paired analysis of the
pretrained temporal prior are reported in
Appendix~\ref{app:per_domain_ablations}.

\section{Conclusion}
\label{sec:conclusion}

We study action-conditioned time-series modeling from the perspective of
world modeling for decision making, where the goal is not only to predict the
realized future but also to reason about outcomes under alternative actions.
We introduce \modelname, which learns action-conditioned dynamics over variable-specific latent states from factual trajectories and rolls the same dynamics forward under candidate future actions.
This formulation supports factual forecasting, intervention-effect prediction,
and downstream planning without requiring paired alternative-action outcomes
or planning supervision.
Across four simulator-grounded controlled systems, \modelname performs strongly
throughout this evaluation pipeline, while our analysis shows that factual
accuracy, intervention fidelity, and decision quality are related but not
interchangeable.
These findings motivate treating alternative-action reasoning and downstream
decision quality as first-class objectives when developing and evaluating
action-conditioned time-series models.

\bibliography{iclr2027}
\bibliographystyle{iclr2027}

\clearpage
\appendix
\appendix

\section{Experimental Details}
\label{app:experimental_details}

\subsection{Environments and Data Generation}
\label{app:data_generation}

\paragraph{Shared protocol.}
All datasets are generated directly from their corresponding simulators.
Supervised windows are constructed independently within sequence boundaries
using stride one, so no window crosses an episode or snippet boundary.
Each sample contains a history of length $T$ and one realized $H$-step future
action and target sequence.
Training and validation combine continuous simulator trajectories with short
factual snippets designed to broaden action coverage.
No paired outcomes under alternative future actions are included in the
factual training data, and the test splits contain continuous trajectories
only.

We use variable-wise z-score normalization with statistics computed from the
training observations only and apply the same statistics to validation and
test data.
Table~\ref{tab:appendix_data_statistics} reports the resulting numbers of
unique supervised windows.

\begin{table}[h]
\centering
\caption{
Factual data used for training and evaluation.
Window counts refer to unique supervised windows.
}
\label{tab:appendix_data_statistics}
\small
\renewcommand{\arraystretch}{1.10}
\setlength{\tabcolsep}{6pt}
\begin{tabular}{lrrrl}
\toprule
Domain
& Train
& Validation
& Test
& Split construction \\
\midrule
CityLearn
& 163,920
& 26,520
& 23,320
& Disjoint calendar ranges \\
SUMO
& 88,000
& 19,400
& 13,000
& Disjoint simulator seeds \\
GreenLight
& 74,444
& 23,428
& 22,596
& Disjoint weather years \\
Inventory
& 35,368
& 8,748
& 7,808
& Disjoint simulator seeds \\
\bottomrule
\end{tabular}
\end{table}

\paragraph{CityLearn.}
We use the CityLearn challenge 2022 phase-1 environment
\citep{nweye2023citylearn} with a single building and hourly control.
The 28-dimensional observation includes calendar information, current weather
and short-horizon weather forecasts, carbon intensity, non-shiftable load,
solar generation, electricity price information, battery state of charge, and
net electricity consumption.
The scalar action controls battery charging and discharging.
We predict battery state of charge and net electricity consumption.

Factual trajectories are generated using a mixture of smooth behavior
control, randomized control, structured action perturbations, and near-zero
control.
The train, validation, and test splits occupy disjoint portions of the
CityLearn time axis, using the first 70\%, the following 15\%, and the final
15\%, respectively.
Training and validation additionally contain 36-step snippets consisting of a
24-step history followed by one assigned 12-step future action sequence.

\paragraph{SUMO.}
We use the RESCO Cologne1 single-intersection scenario in SUMO
\citep{ault2021reinforcement,lopez2018microscopic}.
Each episode covers one hour of traffic with control decisions every
10 seconds.
The 25-dimensional observation summarizes directional queues and vehicle
counts, normalized speeds, waiting time, recent traffic flow, signal state,
demand level, and periodic time features.
The action selects one of four requested green phases and is represented as a
four-dimensional one-hot vector.
The six prediction targets comprise four directional queue counts, aggregate
waiting time, and network arrivals.

Factual trajectories combine adaptive pressure-based control, perturbed
adaptive control, jittered fixed-time control, and randomized phase blocks,
with traffic demand varied across episodes.
Train, validation, and test use disjoint simulator-seed streams while sharing
the same Cologne1 road network and traffic scenario.
Training and validation also include factual snippets formed from a
behavior-generated prefix followed by one assigned phase-control plan.

\paragraph{GreenLight.}
We use the GreenLight tomato greenhouse environment
\citep{van2025greenlight,katzin2020greenlight} with a 15-minute control
interval and Amsterdam weather.
The 36-dimensional observation combines greenhouse state, current outdoor
weather, time features, and future-known weather summaries over the next four
hours.
The six actions control heating, CO$_2$ injection, thermal screening,
ventilation, lighting, and blackout screening, with all controls represented
as absolute levels in $[0,1]$.
We predict indoor CO$_2$, air temperature, and relative humidity.

Factual trajectories are generated from rule-based climate control with
persistent perturbations and structured control blocks.
The split is defined by weather year, using 2008--2010 for training, 2011 for
validation, and 2012 for testing.
Within each year, simulations cover four non-overlapping 60-day seasons.
Training and validation additionally include snippets sampled from separately
generated seasonal trajectories.
Each snippet contains a 96-step history followed by one 16-step control
sequence drawn from reference and actuator-specific low- and high-control
plans.

\paragraph{Inventory management.}
We use the serial environment from \texttt{gym-invmgmt}
\citep{barati2026gym} with backlog enabled.
The system consists of retailer, distributor, and factory stages and evolves
over 96 decision periods per episode.
The 15-dimensional observation contains recent customer demand, customer
backlog, inventories at the three stages, pipeline arrivals, the current
period, and customer sentiment.
The three actions specify requested replenishment quantities for the retailer,
distributor, and factory.
Physical requests are bounded between 0 and 60 units with a maximum
period-to-period change of 15 units and are stored after scaling to $[0,1]$.
The prediction targets are customer backlog and the inventories of the three
stages.

Demand conditions vary across episodes in their mean, seasonality, trend, and
stochastic variation.
Factual behavior is generated by causal base-stock control augmented with
smooth residual variation and structured order perturbations.
Train, validation, and test use disjoint simulator-seed streams.
Training and validation additionally contain snippets consisting of a
24-step history followed by one assigned 12-step replenishment plan.
The plan family varies the magnitude and timing of replenishment changes
across individual channels and coordinated system-wide actions.

\subsection{Evaluation Protocols and Metrics}
\label{app:evaluation}

We evaluate all methods in four settings: factual prediction,
intervention-effect prediction, offline decision making, and closed-loop
planning.
For intervention and offline evaluation, all methods use the same held-out
histories, candidate action banks, and simulator-generated outcomes.
For each held-out history, all candidate action sequences are rolled out from the identical simulator state at the history boundary. Stochastic simulator components and exogenous realizations are matched across candidates, so differences in future trajectories arise only from the candidate actions.
Alternative-action outcomes are used only for evaluation and never for
training or model selection.
Within each domain, closed-loop evaluation uses the same simulator episodes,
candidate-generation protocol, and CEM budget across learned methods.
Table~\ref{tab:appendix_eval_protocol} summarizes the evaluation scale.

\begin{table}[h]
\centering
\caption{
Evaluation scale across domains.
Intervention and Offline report the number of histories and candidate plans per history.
For online planning, $K$ denotes the candidate population size per CEM round
and $H$ the planning horizon.
}
\label{tab:appendix_eval_protocol}
\small
\renewcommand{\arraystretch}{1.10}
\setlength{\tabcolsep}{5pt}
\begin{tabular}{lcccc}
\toprule
Domain
& Intervention histories $\times$ plans
& Offline histories $\times$ plans
& Online episodes
& Online $K/H$ \\
\midrule
CityLearn  & $100\times16$ & $512\times64$ & 8  & $64/12$ \\
SUMO       & $96\times6$   & $256\times6$  & 20 & $16/12$ \\
GreenLight & $256\times13$ & $384\times16$ & 24 & $24/16$ \\
Inventory  & $256\times9$  & $480\times24$ & 24 & $24/12$ \\
\bottomrule
\end{tabular}
\end{table}

\paragraph{Factual prediction.}
Factual errors are measured in target-wise standardized space using statistics
computed from the factual training data.
For target $p$, define
\[
e_{ihp}
=
\frac{\hat y_{ihp}-y_{ihp}}{\sigma_p},
\]
where $i$, $h$, and $p$ index the test window, prediction horizon, and target.
We report
\[
\mathrm{MAE}
=
\operatorname{mean}_{i,h,p}|e_{ihp}|,
\qquad
\mathrm{MSE}
=
\operatorname{mean}_{i,h,p}e_{ihp}^{2}.
\]
The main paper reports MAE, while the detailed results include both metrics.
For \modelname{}, the $0.5$ quantile is used as the point prediction.

\paragraph{Intervention-effect prediction.}
For each held-out history, candidate $k=0$ is a reference plan and $k>0$
indexes alternative future action sequences.
The true and predicted intervention effects are defined relative to the same
reference trajectory,
\[
\Delta y_{ikhp}
=
y^{(k)}_{ihp}-y^{(0)}_{ihp},
\qquad
\Delta\hat y_{ikhp}
=
\hat y^{(k)}_{ihp}-\hat y^{(0)}_{ihp}.
\]
We evaluate effect fidelity using
\[
E_{\mathrm{err}}
=
\operatorname{mean}_{i,k>0,h,p}
\left|
\frac{\Delta\hat y_{ikhp}-\Delta y_{ikhp}}{\sigma_p}
\right|,
\]
\[
E_{\mathrm{true}}
=
\operatorname{mean}_{i,k>0,h,p}
\left|
\frac{\Delta y_{ikhp}}{\sigma_p}
\right|,
\qquad
\mathrm{Effect\ NMAE}
=
\frac{E_{\mathrm{err}}}{E_{\mathrm{true}}+10^{-12}}.
\]
The reference plan is excluded from the average.
Lower Effect NMAE indicates more accurate prediction of action-induced
trajectory changes.

\paragraph{Offline decision making.}
Offline evaluation measures whether predicted trajectories support reliable
selection among candidate action plans.
For history $i$, objective $j$, and candidate $k$, let
$\widehat C_{ijk}$ and $C_{ijk}$ denote the costs computed from the predicted
and simulator trajectories.
The model selects a candidate according to predicted cost under the
domain-specific constraints, while the oracle uses the true simulator costs.
The random reference is the exact average over the eligible candidate bank.

For objective $j$, normalized regret is
\[
R_j
=
\frac{
\frac{1}{N}\sum_i C_{ij,\hat k_i}
-
\frac{1}{N}\sum_i C_{ij,k_i^\star}
}{
\max\left(
\frac{1}{N}\sum_i C^{\mathrm{rand}}_{ij}
-
\frac{1}{N}\sum_i C_{ij,k_i^\star},
10^{-12}
\right)
},
\]
where $\hat k_i$ and $k_i^\star$ are the model-selected and oracle candidates.
The reported Macro Regret is the equal-weight average over three predefined
objectives,
\[
R_{\mathrm{macro}}
=
\frac{1}{3}\sum_{j=1}^{3}R_j.
\]
Lower values indicate decisions closer to the simulator oracle.

\paragraph{Decision objectives.}
The objectives capture complementary operating criteria in each environment
and are fixed across all methods.
Table~\ref{tab:decision_objectives} summarizes the three objectives used for
offline evaluation and the primary objective used in closed-loop planning.
The same definitions are applied to predicted and simulator trajectories.

\begin{table}[h]
\centering
\caption{Decision objectives used in offline and closed-loop evaluation.}
\label{tab:decision_objectives}
\small
\renewcommand{\arraystretch}{1.12}
\setlength{\tabcolsep}{4.5pt}
\begin{tabular}{lll}
\toprule
Domain & Offline objectives & Online objective \\
\midrule
CityLearn
& Electricity cost, grid peak, grid energy
& Electricity cost \\
SUMO
& Balanced congestion, queue, waiting
& Balanced congestion \\
GreenLight
& Mean climate violation, peak climate violation, balanced cost
& Balanced cost \\
Inventory
& Economic stock cost, service risk, recovery cost
& Economic stock cost \\
\bottomrule
\end{tabular}
\end{table}

For CityLearn, the three objectives respectively measure price-weighted grid
import, peak grid import, and average grid import.
Each also includes a terminal SOC recovery penalty to discourage plans that
obtain short-term gains by excessively depleting the battery.
Let $g_h=\max(y_h^{\mathrm{net}},0)$ denote grid import and $p_h$ the
electricity price.
We define
\[
G=\frac{\max(s_0-0.10-s_H,0)}{0.90},
\]
and
\[
C_{\mathrm{cost}}
=
\frac{\operatorname{mean}_h(g_hp_h)}{c_{\mathrm{cost}}}+G,
\qquad
C_{\mathrm{peak}}
=
\frac{\max_h g_h}{c_{\mathrm{peak}}}+G,
\qquad
C_{\mathrm{energy}}
=
\frac{\operatorname{mean}_h g_h}{c_{\mathrm{energy}}}+G,
\]
where
$c_{\mathrm{cost}}=0.2769$,
$c_{\mathrm{peak}}=3.2898$, and
$c_{\mathrm{energy}}=0.9899$
are fixed normalization scales used throughout evaluation.

For SUMO, we consider complementary measures of average congestion and
traffic balance.
Let $q$, $w$, and $f$ denote normalized mean total queue, mean waiting time,
and maximum directional queue,
\[
q=\frac{\operatorname{mean}_h\sum_d Q_{hd}}{40},
\qquad
w=\frac{\operatorname{mean}_h W_h}{1500},
\qquad
f=\frac{\operatorname{mean}_h\max_d Q_{hd}}{20},
\]
and let $s$ denote the mean phase-switch rate.
The balanced objective combines these components as
\[
C_{\mathrm{balanced}}
=
q+0.5w+0.2f+0.03s,
\]
while
\[
C_{\mathrm{queue}}=q,
\qquad
C_{\mathrm{waiting}}=w.
\]
The additional terms in the balanced objective penalize directional
bottlenecks and excessive signal switching beyond average congestion alone.

For GreenLight, we use the admissible climate ranges and violation scales
defined by GL-Gym for indoor CO$_2$, temperature, and relative humidity.
For target $p$, normalized violation is
\[
v_{hp}
=
\frac{\max(L_p-y_{hp},0)+\max(y_{hp}-U_p,0)}{S_p},
\]
with
\[
L=(300,15,50),\qquad
U=(1600,34,85),\qquad
S=(2500,15,15).
\]
We evaluate mean and peak climate violation,
\[
C_{\mathrm{mean}}
=
\operatorname{mean}_h \sum_p v_{hp},
\qquad
C_{\mathrm{peak}}
=
\max_h \sum_p v_{hp}.
\]
The balanced objective additionally accounts for operating effort,
\[
C_{\mathrm{balanced}}
=
\operatorname{mean}_h
\left(
\sum_p v_{hp}+o_h
\right),
\]
where
\[
o_h
=
\frac{
0.002925\,u^{\mathrm{Boil}}_h
+
0.001350\,u^{\mathrm{CO_2}}_h
+
0.008700\,u^{\mathrm{Lamp}}_h
}{
0.02024146
}.
\]
Here, $u^{\mathrm{Boil}}_h$, $u^{\mathrm{CO_2}}_h$, and
$u^{\mathrm{Lamp}}_h$ are actuator fractions, and the coefficients and
normalization follow the cost and normalization conventions of GL-Gym. The balanced objective therefore captures both climate regulation and
operating effort on comparable normalized scales.

For Inventory, backlog and inventory levels are evaluated on their
nonnegative simulator support.
Let $B_h$ denote customer backlog and $I^R_h$, $I^D_h$, and $I^F_h$ the
retailer, distributor, and factory inventories.
Define
\[
g_h
=
0.03 I^R_h + 0.02 I^D_h + 0.01 I^F_h,
\qquad
\ell_h
=
0.1 B_h + g_h.
\]
The economic objective is
\[
C_{\mathrm{economic}}
=
\operatorname{mean}_h \ell_h,
\]
using the backlog and holding-cost coefficients of the serial inventory
environment.

To capture service risk, we emphasize periods with the largest backlog.
For a trajectory of length $n$, let
$m=\max(1,\lceil 0.25n\rceil)$ and
$\operatorname{TopMean}_m(B_{1:n})$ denote the mean of its $m$ largest backlog
values. We define
\[
C_{\mathrm{service}}
=
\operatorname{mean}_h g_h
+
0.1\,\operatorname{TopMean}_m(B_{1:n}).
\]
Finally, recovery quality is measured from the terminal operating state,
\[
C_{\mathrm{recovery}}
=
\frac{1}{r}
\sum_{h=n-r+1}^{n}\ell_h,
\qquad
r=\min(4,n),
\]
where the four-period window matches the replenishment lead time of the serial
environment.
Together, the three objectives measure average operating cost, service risk,
and terminal recovery.

In addition to normalized regret, we evaluate whether the predicted costs
preserve the relative ordering of candidate plans. For each held-out history
and decision objective, we compute the Spearman correlation between the
predicted candidate-cost vector and the corresponding simulator-cost vector.
Higher correlation indicates more faithful candidate ranking. We report this
metric in Section~\ref{sec:ranking_fidelity}.

\paragraph{Closed-loop planning.}
We use receding-horizon model predictive control (MPC) with the cross-entropy
method (CEM) \citep{de2005tutorial} to optimize candidate action sequences.
At each decision step, the model evaluates an initial population of $K$
candidate plans over horizon $H$, after which CEM refines the proposal
distribution using the lowest-cost candidates.
CityLearn uses continuous Gaussian CEM with three refinement rounds and
12 elites.
SUMO uses categorical CEM over feasible traffic-signal phases with two
refinement rounds and four elites.
GreenLight uses categorical CEM over admissible actuator macro-actions, while
Inventory uses continuous CEM over residual replenishment plans; both use two
refinement rounds and six elites.

Candidate generation respects the action constraints of each environment,
including battery feasibility in CityLearn, traffic-signal phase constraints
in SUMO, admissible greenhouse controls in GreenLight, and replenishment bounds
and rate limits in Inventory.
Candidates are ranked using the model-predicted primary objective listed in
Table~\ref{tab:decision_objectives}.
Only the first action of the selected plan is executed in the simulator.
The resulting observation is appended to the history and the planning process
is repeated.
Within each domain, all learned methods use the same candidate-generation
rules and CEM budget.

Online scores are computed from complete realized simulator episodes and then
averaged across episodes.
Lower is better in every domain.
Because the objectives have different physical meanings and scales, online
values are compared only within the same domain.

\paragraph{Cross-level alignment.}
We measure whether model rankings are preserved across adjacent evaluation
levels.
For prediction-to-effect alignment, the per-history prediction score of method
$m$ is the standardized MAE on the reference trajectory,
\[
P_i^{(m)}
=
\operatorname{mean}_{h,p}
\left|
\frac{\hat y^{(m)}_{i0hp}-y_{i0hp}}{\sigma_p}
\right|,
\]
and $E_i^{(m)}$ is the corresponding per-history Effect NMAE over alternative
plans.

For effect-to-decision alignment, the effect score is computed on the offline
candidate bank.
We use the same candidate-cost ranking
criterion under the primary domain objective,
\[
D_i^{(m)}
=
-\rho_S
\left(
\widehat{\mathbf C}^{(m)}_i,
\mathbf C_i
\right),
\]
where $\rho_S$ denotes Spearman rank correlation.
The negative sign makes lower values better for all three scores.

For each history and unordered method pair $(a,b)$, an ordering reversal occurs
when
\[
\bigl(x_i^{(a)}-x_i^{(b)}\bigr)
\bigl(y_i^{(a)}-y_i^{(b)}\bigr)<0,
\]
where $x$ and $y$ denote adjacent evaluation levels.
Tied pairs are excluded.
The domain-level reversal rate is the mean fraction of reversed method pairs
over its histories, and the reported cross-domain value is the equal-weight
mean over the four domains.
We also compute Spearman correlation and Kendall's $\tau_b$ across the eight
methods for each history and aggregate both statistics in the same way.
Confidence intervals in Figure~\ref{fig:metric_alignment} are obtained from
2,000 bootstrap resamples of histories within each domain.


\subsection{Implementation Details}
\label{app:implementation}

\paragraph{Model configuration.}
We use the same architecture and shared hyperparameters across all four
domains, with environment-specific input dimensions and temporal horizons.
Each observed variable is represented by a 144-dimensional latent state.
The controlled dynamics use two transition blocks with four-head
self-attention across variables and a SwiGLU update.
The observation encoder and action router both incorporate first differences,
as described in Section~\ref{sec:state_dynamics}.
The prediction head outputs the $0.1$, $0.5$, and $0.9$ quantiles, with the
median used for point prediction.

\paragraph{Temporal prior.}
We use a frozen six-layer EIDOS backbone with hidden dimension 384 and
12 attention heads.
Only the target histories are passed to EIDOS.
Each target is standardized over its observed history, encoded independently, and passed through the same frozen projection to 128 dimensions.
The projected representations are averaged across targets, and the final
history representation is passed through the residual adapter in
Section~\ref{sec:prior_rollout}.
The adapter gate is initialized to zero.
The EIDOS backbone remains frozen throughout training, and the temporal prior
is injected only at the history boundary.

\paragraph{Training and model selection.}
Stage A trains the controlled core and prediction head for 10 epochs using
Adam with learning rate $10^{-3}$ and batch size 256.
Stage B freezes the Stage-A predictor and the EIDOS backbone and trains only
the 60,321-parameter history-prior adapter for 3 epochs using AdamW with
learning rate $10^{-3}$ and weight decay $10^{-2}$.
Both stages optimize the mean pinball loss \citep{koenker1978regression} over the three predicted quantiles
and use gradient clipping with global norm 1.0.
Final checkpoints are selected solely by factual validation MAE.
No intervention-effect or planning metric is used for model selection.

\subsection{Baseline Adaptations}
\label{app:baselines}

All baselines are trained on the same factual data splits as \modelname{}.
For intervention and offline evaluation, they use the same held-out histories,
candidate action banks, and simulator outcomes.
Historical actions are provided together with the observed history, and future
actions are introduced causally so that prediction at horizon $h$ never
depends on actions beyond that horizon.
Within each domain, closed-loop evaluation uses the same simulator episodes,
candidate-generation protocol, and CEM budget across learned methods.
Baseline checkpoints are selected using factual validation MAE only.

\paragraph{Forecasting baselines.}
We adapt Ridge-VARX, DLinear, PatchTST, iTransformer, and TFT to condition on
future actions while retaining their native forecasting backbones.
Ridge-VARX performs direct multi-horizon regression using recent observations,
historical actions, and the future-action prefix available at each horizon.
Its lag length and ridge coefficient are selected on factual validation data.

For DLinear, PatchTST, and iTransformer, we attach a causal
action-conditioning head to the corresponding time-series encoder.
The representation for each target and forecast horizon is combined with the
historical actions and the available prefix of candidate future actions before
prediction.
DLinear operates on target histories after trend-seasonal decomposition,
PatchTST uses channel-independent temporal representations, and iTransformer
allows information exchange across observation variables before action
conditioning.
TFT follows its native known-future-input formulation.
Historical observations and actions are processed by the historical encoder,
while candidate future actions are supplied to the decoder as known-future
inputs.
Its median quantile is used as the point prediction.

\paragraph{World-model baselines.}
RSSM and MTS3 are evaluated as action-conditioned latent dynamics models.
RSSM infers its latent state from the observed history through
action-conditioned transitions and observation updates, then rolls the latent
prior forward under candidate future actions.
MTS3 similarly filters the observed history into structured latent beliefs and
propagates them under future actions at its fast and slow temporal scales.
For both models, future observations are unavailable during evaluation, and
target trajectories are decoded from the action-conditioned latent rollout.

\section{Detailed Experimental Results}
\label{app:detailed_results}

\subsection{Task-Level Results}

We report the detailed results for the four evaluation settings defined in
Appendix~\ref{app:evaluation}.
In addition to the primary metrics reported in the main paper, we include
complementary metrics for factual prediction, intervention-effect prediction,
and closed-loop planning.
All task metrics reported in this subsection are lower-is-better.
Best results are shown in \textbf{bold} and second-best results are
\underline{underlined}.
When values are identical at the reported precision, tied entries share the
same formatting.

\paragraph{Factual prediction.}

\begin{table*}[h]
    \centering
    \small
    \setlength{\tabcolsep}{4.2pt}
    \caption{
    Factual prediction results across the four domains.
    MAE and MSE are computed in standardized target space.
    }
    \label{tab:appendix-factual}
    \begin{tabular}{lrrrrrrrr}
        \toprule
        Method
        & \multicolumn{2}{c}{CityLearn}
        & \multicolumn{2}{c}{SUMO}
        & \multicolumn{2}{c}{GreenLight}
        & \multicolumn{2}{c}{Inventory} \\
        \cmidrule(lr){2-3}
        \cmidrule(lr){4-5}
        \cmidrule(lr){6-7}
        \cmidrule(lr){8-9}
        & MAE $\downarrow$ & MSE $\downarrow$
        & MAE $\downarrow$ & MSE $\downarrow$
        & MAE $\downarrow$ & MSE $\downarrow$
        & MAE $\downarrow$ & MSE $\downarrow$ \\
        \midrule

        Ridge-VARX
        & 0.2875 & 0.2003
        & 0.3873 & 0.3059
        & 0.3263 & 0.1873
        & 0.2210 & 0.1169 \\

        DLinear-Action
        & 0.3585 & 0.2667
        & 0.4358 & 0.4540
        & 0.3699 & 0.2575
        & 0.3812 & 0.3169 \\

        PatchTST-Action
        & 0.3410 & 0.2589
        & 0.4150 & 0.4051
        & 0.3647 & 0.2485
        & 0.3938 & 0.3314 \\

        iTransformer-Action
        & 0.3602 & 0.2749
        & 0.3730 & 0.3287
        & 0.3398 & 0.2173
        & 0.3706 & 0.2995 \\

        TFT-Action
        & 0.2583 & 0.2354
        & \underline{0.1226} & \underline{0.0521}
        & \underline{0.0666} & \underline{0.0099}
        & \underline{0.0992} & \underline{0.0391} \\

        RSSM
        & \underline{0.2501} & \textbf{0.1846}
        & 0.1730 & 0.0979
        & 0.1008 & 0.0210
        & 0.1243 & 0.0626 \\

        MTS3
        & 0.2512 & \underline{0.1945}
        & 0.2134 & 0.1198
        & 0.1242 & 0.0323
        & 0.1279 & 0.0536 \\

        \midrule
        \modelname{}
        & \textbf{0.2486} & 0.2017
        & \textbf{0.1076} & \textbf{0.0403}
        & \textbf{0.0498} & \textbf{0.0057}
        & \textbf{0.0790} & \textbf{0.0300} \\

        \bottomrule
    \end{tabular}
\end{table*}

Table~\ref{tab:appendix-factual} reports factual prediction performance using
both MAE and MSE in standardized target space.
\modelname{} achieves the lowest MAE in all four domains and the lowest MSE on
SUMO, GreenLight, and Inventory.

\paragraph{Intervention-effect prediction.}

\begin{table*}[t]
    \centering
    \small
    \caption{
    Intervention-effect prediction results across the four domains.
    Effect NMAE pools normalized intervention-effect errors across targets,
    while Macro NMAE averages target-wise normalized effect errors.
    }
    \label{tab:appendix-counterfactual}
    \resizebox{\textwidth}{!}{%
    \begin{tabular}{lrrrrrrrr}
        \toprule
        Method
        & \multicolumn{2}{c}{CityLearn}
        & \multicolumn{2}{c}{SUMO}
        & \multicolumn{2}{c}{GreenLight}
        & \multicolumn{2}{c}{Inventory} \\
        \cmidrule(lr){2-3}
        \cmidrule(lr){4-5}
        \cmidrule(lr){6-7}
        \cmidrule(lr){8-9}

        & Effect NMAE $\downarrow$
        & Macro NMAE $\downarrow$
        & Effect NMAE $\downarrow$
        & Macro NMAE $\downarrow$
        & Effect NMAE $\downarrow$
        & Macro NMAE $\downarrow$
        & Effect NMAE $\downarrow$
        & Macro NMAE $\downarrow$ \\
        \midrule

        Ridge-VARX
        & 0.1495 & 0.1646
        & 1.0851 & 1.1003
        & 0.8622 & 0.8624
        & 0.6538 & 1.0046 \\

        DLinear-Action
        & 0.3493 & 0.3656
        & 0.9201 & 0.9259
        & 0.9384 & 0.9479
        & 0.9711 & 1.1413 \\

        PatchTST-Action
        & 0.2700 & 0.2730
        & 0.9202 & 0.9266
        & 0.9221 & 0.9368
        & 0.9774 & 1.1116 \\

        iTransformer-Action
        & 0.3972 & 0.3943
        & 0.9060 & 0.9118
        & 0.8860 & 0.8835
        & 0.9567 & 1.0885 \\

        TFT-Action
        & \textbf{0.0498} & \textbf{0.0718}
        & \textbf{0.4184} & \textbf{0.4213}
        & \underline{0.2811} & \underline{0.3059}
        & \underline{0.3077} & \underline{0.6023} \\

        RSSM
        & 0.0790 & 0.1214
        & 0.4914 & 0.4965
        & 0.3761 & 0.4178
        & 0.3506 & 0.6604 \\

        MTS3
        & \underline{0.0716} & 0.1137
        & 0.5776 & 0.5820
        & 0.5367 & 0.5725
        & 0.4008 & 0.7660 \\

        \midrule
        \modelname{}
        & 0.0725 & \underline{0.1075}
        & \underline{0.4825} & \underline{0.4751}
        & \textbf{0.2091} & \textbf{0.2434}
        & \textbf{0.2244} & \textbf{0.4645} \\

        \bottomrule
    \end{tabular}
    }
\end{table*}

Table~\ref{tab:appendix-counterfactual} reports pooled Effect NMAE together
with Macro NMAE, which averages target-wise normalized effect errors.
\modelname{} achieves the lowest error under both metrics on GreenLight and
Inventory and the second-lowest error on SUMO.
The pooled and target-balanced metrics show broadly consistent performance
patterns across methods.

\paragraph{Offline decision making.}

\begin{table*}[h]
    \centering
    \small
    \setlength{\tabcolsep}{9pt}
    \caption{
    Offline decision results on fixed candidate-action banks.
    Macro Regret is the equal-weight average of normalized regret over the
    three decision objectives defined in Appendix~\ref{app:evaluation}.
    }
    \label{tab:appendix-offline}

    \begin{tabular}{lrrrr}
        \toprule
        Method
        & CityLearn
        & SUMO
        & GreenLight
        & Inventory \\
        \midrule
        & \multicolumn{1}{c}{Macro Regret $\downarrow$}
        & \multicolumn{1}{c}{Macro Regret $\downarrow$}
        & \multicolumn{1}{c}{Macro Regret $\downarrow$}
        & \multicolumn{1}{c}{Macro Regret $\downarrow$} \\
        \midrule

        Ridge-VARX
        & 0.4793
        & 0.5255
        & 0.4352
        & \textbf{0.1169} \\
        
        DLinear-Action
        & 0.3869
        & 0.5661
        & 0.7962
        & 0.2929 \\
        
        PatchTST-Action
        & 0.4396
        & 0.5662
        & 0.8965
        & 0.3221 \\
        
        iTransformer-Action
        & 0.5050
        & 0.5620
        & 0.7096
        & 0.3133 \\
        
        TFT-Action
        & \underline{0.2880}
        & \underline{0.0296}
        & 0.2367
        & 0.1585 \\
        
        RSSM
        & 0.3739
        & 0.0802
        & 0.1956
        & 0.2130 \\
        
        MTS3
        & 0.3105
        & 0.0759
        & \underline{0.1870}
        & \underline{0.1531} \\
        
        \midrule
        \modelname{}
        & \textbf{0.2798}
        & \textbf{0.0221}
        & \textbf{0.1397}
        & 0.1563 \\
        \bottomrule
    \end{tabular}
\end{table*}

Table~\ref{tab:appendix-offline} reports Macro Regret on the fixed
candidate-action banks.
\modelname{} achieves the lowest regret on CityLearn, SUMO, and GreenLight,
giving the best result in three of the four domains.

\paragraph{Closed-loop planning.}

\begin{table*}[h]
    \centering
    \caption{
    Closed-loop planning results across the four domains.
    The first metric for each domain is the primary objective reported in the
    main paper, followed by two complementary operating metrics.
    Scores are comparable across methods within a domain but not across
    domains.
    }
    \label{tab:appendix_online}
    \small
    \setlength{\tabcolsep}{2.8pt}

    \resizebox{\textwidth}{!}{%
    \begin{tabular}{lcccccccccccc}
        \toprule
        Method
        & \multicolumn{3}{c}{CityLearn}
        & \multicolumn{3}{c}{SUMO}
        & \multicolumn{3}{c}{GreenLight}
        & \multicolumn{3}{c}{Inventory} \\
        \cmidrule(lr){2-4}
        \cmidrule(lr){5-7}
        \cmidrule(lr){8-10}
        \cmidrule(lr){11-13}

        & \shortstack{Electricity\\Cost $\downarrow$}
        & \shortstack{Grid\\Energy $\downarrow$}
        & \shortstack{Grid\\Peak $\downarrow$}

        & \shortstack{Balanced\\Congestion $\downarrow$}
        & \shortstack{Queue $\downarrow$}
        & \shortstack{Stopped\\Waiting $\downarrow$}

        & \shortstack{Balanced\\Cost $\downarrow$}
        & \shortstack{Mean Climate\\Violation $\downarrow$}
        & \shortstack{Peak Climate\\Violation $\downarrow$}

        & \shortstack{Economic\\Cost $\downarrow$}
        & \shortstack{Service-risk\\Cost $\downarrow$}
        & \shortstack{Recovery\\Cost $\downarrow$} \\
        \midrule

        Ridge-VARX
        & 0.8310
        & 0.8053
        & \underline{1.0737}
        & 0.3940
        & 0.2689
        & 0.1105
        & 0.3105
        & 0.0363
        & 0.3818
        & 4.108
        & 4.402
        & 3.669 \\

        DLinear-Action
        & 0.8541
        & 0.8313
        & 1.2742
        & 0.3834
        & 0.2601
        & 0.8416
        & 0.3201
        & 0.0376
        & 0.3995
        & 4.295
        & 6.196
        & 4.107 \\

        PatchTST-Action
        & 0.8481
        & 0.8250
        & 1.2708
        & 0.3836
        & 0.2451
        & 0.8384
        & 0.3232
        & 0.0401
        & 0.4444
        & 4.376
        & 6.461
        & 4.390 \\

        iTransformer-Action
        & 0.9099
        & 0.8722
        & 1.3994
        & 0.3843
        & 0.2518
        & 0.8329
        & 0.3258
        & 0.0380
        & 0.3829
        & 4.288
        & 5.821
        & 4.046 \\

        TFT-Action
        & \underline{0.8250}
        & \underline{0.8019}
        & 1.1022
        & \underline{0.3147}
        & 0.2657
        & 0.0824
        & \underline{0.2946}
        & \textbf{0.0250}
        & \textbf{0.2674}
        & \underline{3.892}
        & \textbf{4.247}
        & 3.868 \\

        RSSM
        & \textbf{0.8196}
        & \textbf{0.7950}
        & \textbf{1.0639}
        & 0.3193
        & \underline{0.2024}
        & \underline{0.0778}
        & 0.3009
        & 0.0349
        & 0.3019
        & 3.958
        & 4.348
        & 3.748 \\

        MTS3
        & 0.8414
        & 0.8131
        & 1.0828
        & 0.3430
        & 0.2188
        & 0.0817
        & 0.2961
        & \underline{0.0300}
        & 0.3094
        & 4.044
        & 4.450
        & \textbf{3.495} \\

        \midrule
        \modelname{}
        & 0.8317
        & 0.8054
        & 1.1220
        & \textbf{0.3036}
        & \textbf{0.1981}
        & \textbf{0.0777}
        & \textbf{0.2944}
        & \underline{0.0300}
        & \underline{0.3010}
        & \textbf{3.874}
        & \underline{4.318}
        & \underline{3.632} \\

        \bottomrule
    \end{tabular}
    }
\end{table*}

Table~\ref{tab:appendix_online} reports the primary closed-loop objective
together with two complementary operating metrics for each domain.
\modelname{} achieves the best primary objective on SUMO, GreenLight, and
Inventory.
On SUMO, it also obtains the lowest queue and stopped-waiting costs, while the
additional GreenLight and Inventory metrics remain competitive.

\subsection{Cross-Level Alignment}
\label{app:cross_level_alignment}

\begin{table}[t]
    \centering
    \caption{
    Alignment between adjacent evaluation levels, macro-averaged across the
    four domains.
    Higher rank correlation and lower reversal rate indicate stronger
    agreement in model ordering.
    }
    \label{tab:metric_alignment}
    \small
    \setlength{\tabcolsep}{6pt}
    \begin{tabular}{lccc}
        \toprule
        Transition
        & Spearman $\rho$
        & Kendall $\tau_b$
        & Reversals $\downarrow$ \\
        \midrule
        Prediction $\rightarrow$ Effect
        & 0.763
        & 0.643
        & 17.8\% \\

        Effect $\rightarrow$ Decision
        & 0.514
        & 0.412
        & 28.4\% \\
        \bottomrule
    \end{tabular}
\end{table}

Table~\ref{tab:metric_alignment} summarizes the alignment between adjacent
evaluation levels.
Model rankings under prediction and intervention fidelity remain positively
associated, with a
macro-averaged Spearman correlation of $0.763$ and Kendall $\tau_b$ of
$0.643$.
However, $17.8\%$ of method pairs reverse their ordering between the two
levels.
The alignment becomes weaker from intervention fidelity to decision quality:
Spearman correlation drops to $0.514$, Kendall $\tau_b$ to $0.412$, and the
reversal rate rises to $28.4\%$.
Thus, although adjacent evaluation levels are related, their model rankings are
not consistently preserved.

\begin{figure*}[h]
    \centering

    \begin{subfigure}{0.45\textwidth}
        \centering
        \includegraphics[
            width=\textwidth
        ]{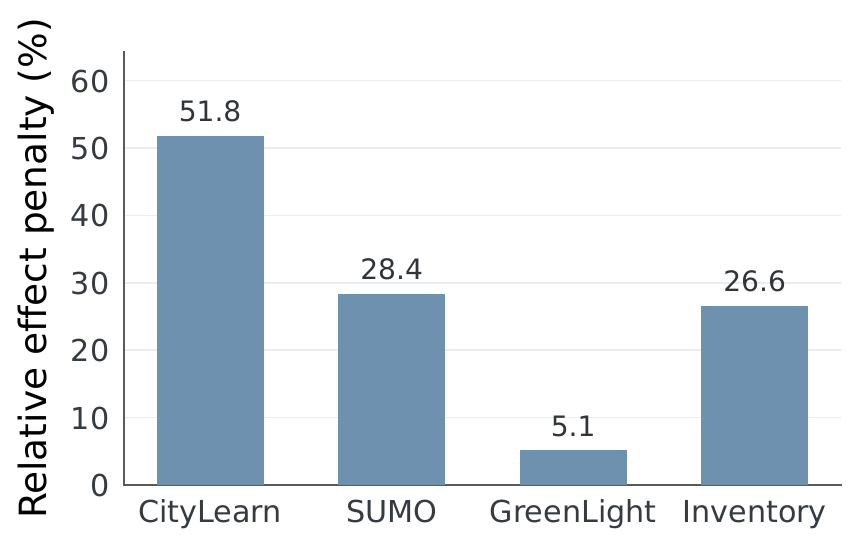}
        \caption{
        Relative increase in Effect NMAE when selecting the
        prediction-best method instead of the effect-best method.
        }
        \label{fig:gap_effect_penalty}
    \end{subfigure}
    \hfill
    \begin{subfigure}{0.45\textwidth}
        \centering
        \includegraphics[
            width=\textwidth
        ]{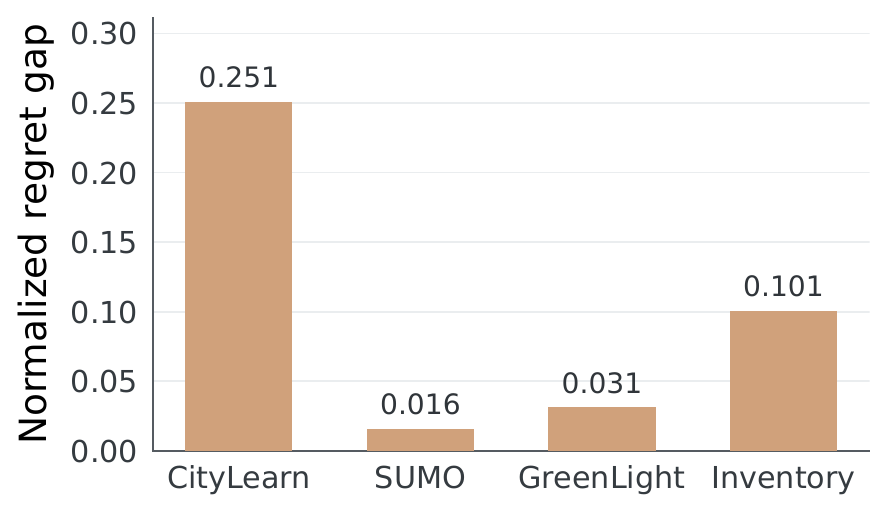}
        \caption{
        Increase in normalized regret when selecting the
        effect-best method instead of the regret-best method.
        }
        \label{fig:gap_regret_penalty}
    \end{subfigure}

    \caption{
    \textbf{Downstream penalty from selecting the best method at the preceding
    evaluation level.}
    For each test history, panel~(a) compares the Effect NMAE of the
    prediction-best method with that of the effect-best method, while
    panel~(b) compares the normalized regret of the effect-best method with
    that of the regret-best method.
    Method selection in this analysis uses the realized test outcome and is
    therefore intended only as a diagnostic of cross-level alignment.
    }
    \label{fig:gap_selection_penalty}
\end{figure*}

Figure~\ref{fig:gap_selection_penalty} examines the practical consequence of
these ranking mismatches.
For each test history, we first identify the method with the lowest upstream
error and measure how much performance is lost at the next evaluation level
relative to the method that is best there.
This is an oracle diagnostic based on realized test outcomes rather than a
model-selection procedure.

When selecting the prediction-best method, the resulting Effect NMAE is higher
than that of the effect-best method by $51.8\%$ on CityLearn, $28.4\%$ on
SUMO, $5.1\%$ on GreenLight, and $26.6\%$ on Inventory
(Figure~\ref{fig:gap_effect_penalty}).
A similar mismatch appears between intervention fidelity and decision quality.
Selecting the effect-best method instead of the regret-best method increases
normalized regret by $0.251$, $0.016$, $0.031$, and $0.101$ on the four
domains, respectively (Figure~\ref{fig:gap_regret_penalty}).
The magnitude varies across environments, but neither upstream metric is a
reliable substitute for evaluating the downstream criterion directly.

\subsection{Candidate-Action Ranking Fidelity}
\label{sec:ranking_appendix}

To examine whether the ranking results depend on a particular decision
criterion, we extend the analysis to all three offline objectives in each
domain, yielding 12 domain-objective combinations. In addition to Spearman
correlation, we report Kendall's $\tau_b$ and pairwise ordering accuracy, where the latter is the fraction of candidate pairs whose predicted ordering
agrees with the simulator ordering, excluding pairs tied under the simulator
cost.
For each metric, scores are first averaged over held-out histories for each
objective and then macro-averaged across the 12 objectives.

\begin{table}[h]
    \centering
    \caption{
    Candidate-action ranking fidelity across decision objectives.
    Results are macro-averaged over the 12 domain-objective combinations.
    Higher is better. Best and second-best results are shown in bold and
    underlined, respectively.
    }
    \label{tab:ranking_all_objectives}
    \small
    \setlength{\tabcolsep}{4.5pt}
    \begin{tabular}{lcccccccc}
        \toprule
        Metric
        & Ridge
        & DLinear
        & PatchTST
        & iTransformer
        & TFT
        & RSSM
        & MTS3
        & \modelname{} \\
        \midrule
        Spearman $\rho$
        & 0.6103
        & 0.4985
        & 0.4838
        & 0.5117
        & \underline{0.7857}
        & 0.7581
        & 0.7620
        & \textbf{0.8270} \\

        Kendall $\tau_b$
        & 0.5105
        & 0.4025
        & 0.3890
        & 0.4129
        & \underline{0.6817}
        & 0.6486
        & 0.6522
        & \textbf{0.7272} \\

        Pairwise accuracy
        & 0.7472
        & 0.6881
        & 0.6807
        & 0.6913
        & \underline{0.8361}
        & 0.8213
        & 0.8233
        & \textbf{0.8624} \\
        \bottomrule
    \end{tabular}
\end{table}

Table~\ref{tab:ranking_all_objectives} shows a consistent advantage for
\modelname{} across the three ranking criteria. It achieves the highest
macro-average score under Spearman correlation, Kendall's $\tau_b$, and
pairwise ordering accuracy, with 0.8270, 0.7272, and 0.8624, respectively.
Across the 12 individual objectives, \modelname{} ranks first on 11 under
Spearman correlation and on 10 under each of the other two metrics.

These results indicate that the ranking advantage is not tied to the primary
objective used for closed-loop evaluation or to a particular rank statistic.
Across the decision criteria considered here, \modelname{} more consistently
preserves the relative ordering of candidate action plans.

\subsection{Per-Domain Ablations and Temporal Prior Analysis}
\label{app:per_domain_ablations}

\begin{table*}[t]
    \centering
    \caption{
    Per-domain absolute results for the controlled-dynamics ablations
    summarized in Table~\ref{tab:ablation}.
    Full denotes the matched reference from the same controlled-ablation suite.
    Offline results report Macro Regret.
    Online objectives are domain-specific and are therefore comparable only
    across variants within the same domain.
    All metrics are lower-is-better.
    }
    \label{tab:ablation_absolute}
    \small
    \setlength{\tabcolsep}{5pt}
    \resizebox{\textwidth}{!}{%
    \begin{tabular}{llcccc}
        \toprule
        Domain
        & Variant
        & Factual MAE $\downarrow$
        & Effect NMAE $\downarrow$
        & Macro Regret $\downarrow$
        & Online Cost $\downarrow$ \\
        \midrule

        CityLearn
        & Full
        & 0.2510 & 0.0688 & 0.2931 & 0.8323 \\
        & Simple Action Concatenation
        & 0.2518 & 0.0856 & 0.2808 & 0.8394 \\
        & w/o State Correction
        & 0.3260 & 0.1516 & 0.3547 & 0.8565 \\
        & Shared State Across Variables
        & 0.6957 & 0.6521 & 0.4984 & 0.8730 \\
        & w/o Action Conditioning
        & 0.4457 & 1.0000 & 1.9871 & 0.8667 \\

        \midrule

        SUMO
        & Full
        & 0.1107 & 0.5154 & 0.0247 & 0.3024 \\
        & Simple Action Concatenation
        & 0.1125 & 0.3526 & 0.0246 & 0.3210 \\
        & w/o State Correction
        & 0.1971 & 0.4610 & 0.0380 & 0.3161 \\
        & Shared State Across Variables
        & 0.5762 & 1.1126 & 0.1622 & 0.3664 \\
        & w/o Action Conditioning
        & 0.2830 & 1.0000 & 1.0385 & 2.5007 \\

        \midrule

        GreenLight
        & Full
        & 0.0527 & 0.2164 & 0.1090 & 0.2948 \\
        & Simple Action Concatenation
        & 0.0620 & 0.2529 & 0.1778 & 0.2961 \\
        & w/o State Correction
        & 0.2054 & 0.8047 & 0.5123 & 0.3013 \\
        & Shared State Across Variables
        & 0.4060 & 0.8734 & 0.6490 & 0.3676 \\
        & w/o Action Conditioning
        & 0.3952 & 1.0000 & 1.1305 & 0.4297 \\

        \midrule

        Inventory
        & Full
        & 0.0800 & 0.2240 & 0.1518 & 3.8884 \\
        & Simple Action Concatenation
        & 0.0834 & 0.2336 & 0.1466 & 3.9168 \\
        & w/o State Correction
        & 0.1256 & 0.3832 & 0.2145 & 3.9983 \\
        & Shared State Across Variables
        & 0.5977 & 1.2963 & 0.4891 & 4.1798 \\
        & w/o Action Conditioning
        & 0.4264 & 1.0000 & 0.9424 & 4.7649 \\

        \bottomrule
    \end{tabular}
    }
\end{table*}

Table~\ref{tab:ablation_absolute} reports the absolute results underlying the
controlled-dynamics ablations in Table~\ref{tab:ablation}.
These structural variants use the matched reference from the same
controlled-ablation suite, whereas Table~\ref{tab:main_results} and
Table~\ref{tab:temporal_prior_ablation} use the final selected checkpoints.
The per-domain results show that the main trends in the aggregate ablation are
not driven by a single environment.

Collapsing the variable-specific latent states into a shared state causes
one of the largest degradations across the ablations.
This substantially increases factual error, intervention-effect error, and
offline regret in every domain, indicating that retaining separate latent
states for individual observed variables is important for modeling controlled
multivariate dynamics.
Removing action conditioning is similarly detrimental.
Its Effect NMAE is exactly $1.0$ in all four domains, corresponding to the
zero-effect baseline induced when predictions cannot change with the candidate
action sequence.
The degradation also propagates to offline and closed-loop decision quality.

Observation correction primarily affects the quality of the inferred system
state.
Removing it consistently increases factual MAE and offline regret across all
four domains, with particularly large degradation on GreenLight and Inventory.
Likewise, replacing the structured action-conditioned transition with simple
action concatenation is competitive on some individual metrics but is less
consistent across evaluation levels.
The full transition provides a better overall balance between factual
prediction, intervention fidelity, and downstream decision quality.

\paragraph{Temporal prior.}
The pretrained temporal prior is introduced only after the controlled dynamics
have been learned in Stage A.
We therefore evaluate it separately using an exactly paired Stage-A/Stage-B
comparison, where the two variants share identical controlled-dynamics
parameters and differ only in the lightweight history-prior adapter.

\begin{table*}[t]
    \centering
    \caption{
    Paired analysis of the pretrained temporal prior.
    Each entry reports Full / w/o EIDOS Prior.
    The two variants share the same Stage-A controlled core and differ only in
    the Stage-B prior adapter.
    Lower is better.
    }
    \label{tab:temporal_prior_ablation}
    \small
    \setlength{\tabcolsep}{7pt}
    \begin{tabular}{lcccc}
        \toprule
        Domain
        & Factual MAE $\downarrow$
        & Effect NMAE $\downarrow$
        & Macro Regret $\downarrow$
        & Online Cost $\downarrow$ \\
        \midrule
        CityLearn
        & 0.2486 / 0.2476
        & 0.0725 / 0.0703
        & 0.2798 / 0.2784
        & 0.8317 / 0.8335 \\

        SUMO
        & 0.1076 / 0.1100
        & 0.4825 / 0.4355
        & 0.0221 / 0.0207
        & 0.3036 / 0.3097 \\

        GreenLight
        & 0.0498 / 0.0544
        & 0.2091 / 0.2158
        & 0.1397 / 0.1440
        & 0.2944 / 0.2955 \\

        Inventory
        & 0.0790 / 0.0811
        & 0.2244 / 0.2248
        & 0.1563 / 0.1523
        & 3.8739 / 3.8945 \\
        \bottomrule
    \end{tabular}
\end{table*}

Table~\ref{tab:temporal_prior_ablation} shows that adding the temporal prior
improves factual prediction on SUMO, GreenLight, and Inventory and yields lower
closed-loop objective values in all four domains.
Its effect on intervention fidelity and offline regret is more domain dependent.
These results are consistent with the temporal prior serving as a complementary
refinement of the terminal history state, while alternative-action responses
are modeled by the learned action-conditioned dynamics.

\section{Limitations}
\label{app:limitations}

Our evaluation is simulator-grounded.
This is intentional, as evaluating alternative action sequences from the same
history requires access to outcomes that are generally unavailable in
observational time-series data.
The four environments span building energy, traffic control, greenhouse
climate, and inventory management, but performance in these simulators does
not by itself establish transfer to real-world systems.
An important next step is evaluation on real-world controlled-intervention
datasets or deployment settings in which responses to selected actions can
be observed prospectively.
The evaluation therefore measures predictive fidelity under controlled
interventions rather than causal identification from observational data
with unobserved confounding.

\modelname{} is trained exclusively on factual trajectories, without paired
outcomes for alternative actions.
Consequently, learning how outcomes change with action relies on variation
present across the training trajectories.
We use multiple behavior policies and structured control snippets to
broaden this coverage, and evaluate within the feasible action ranges of each
environment.
The present experiments do not address extrapolation to substantially
different action regimes that are unsupported by the training data.
Characterizing uncertainty under such distribution shifts and adapting the
dynamics with limited new interaction are useful directions for future work.

Finally, our planning experiments use finite-horizon MPC with CEM-based
action-sequence optimization.
This setting isolates whether the learned dynamics are accurate enough to
compare candidate actions and support closed-loop decisions, but does not
address substantially longer-horizon planning or hierarchical control.
In addition, although \modelname{} predicts multiple quantiles, the current
planner primarily uses point trajectory predictions when comparing candidate
plans.
Using predictive uncertainty explicitly for risk-sensitive planning is a
natural extension.

\end{document}